\documentclass[10pt,conference,a4paper]{IEEEtran}
\IEEEoverridecommandlockouts
\pdfoutput=1

\usepackage[utf8]{inputenc}
\usepackage{microtype}
\usepackage{graphicx}
\usepackage{subfigure}
\usepackage{booktabs} 
\usepackage{amsmath}
\usepackage{amsfonts}
\usepackage{amssymb}
\usepackage{amsthm}
\usepackage{color}
\usepackage{array}
\usepackage{epstopdf}
\usepackage{epsfig}
\usepackage{footmisc}
\usepackage{amsbsy}
\usepackage{bm}
\usepackage{cases}
\usepackage{multirow}
\usepackage{multicol,lipsum}
\usepackage{comment}
\usepackage{times}
\usepackage{paralist}
\usepackage{wrapfig}
\usepackage{algorithmic}
\usepackage[ruled]{algorithm2e}

\def\BibTeX{{\rm B\kern-.05em{\sc i\kern-.025em b}\kern-.08em
    T\kern-.1667em\lower.7ex\hbox{E}\kern-.125emX}}

\newcommand{\note}[1]{[\textcolor{red}{\textit{#1}}]}

\newcommand{\bfe}{{\mathbf e}}

\newcommand{\bmx}{{\bm x}}

\newcommand{\bmh}{{\bm h}}
\newcommand{\bms}{{\bm s}}

\newcommand{\bme}{{\bm e}}

\newcommand{\bmu}{{\bm u}}
\newcommand{\bmb}{{\bm b}}

\newcommand{\bfW}{\mathbf{W}}

\newcommand{\bmtheta}{\bm{\theta}}

\newcommand{\bmsigma}{\bm{\sigma}}
\newcommand{\bmvartheta}{\bm{\vartheta}}

\newcommand{\set}[1]{\ensuremath{\mathcal #1}}
\newcommand{\grad}[1]{\nabla #1}

\newcommand{\overra}{\overrightarrow}
\newcommand{\overla}{\overleftarrow}

\newcommand\VOWEL{{VOWEL}}
\newcommand\DECOLLE{{DECOLLE}}

\begin{document}
\title{VOWEL: A Local Online Learning Rule for Recurrent Networks of Probabilistic Spiking Winner-Take-All Circuits
}

\author{\IEEEauthorblockN{Hyeryung Jang, Nicolas Skatchkovsky and Osvaldo Simeone}
\IEEEauthorblockA{KCLIP lab, Department of Engineering \\
King's College London, London, United Kingdom\\
\{hyeryung.jang, nicolas.skatchkovsky, osvaldo.simeone\}@kcl.ac.uk}%
\thanks{The authors have received funding from the European Research Council (ERC) under the European Union’s Horizon 2020 Research and Innovation Programme (Grant Agreement No. 725731).}
\\[-3.0ex]
}


\maketitle

\begin{abstract}
Networks of spiking neurons and Winner-Take-All spiking circuits (WTA-SNNs) can detect information encoded in spatio-temporal multi-valued events. These are described by the timing of events of interest, e.g., clicks, as well as by categorical numerical values assigned to each event, e.g., like or dislike. 
Other use cases include object recognition from data collected by neuromorphic cameras, which produce, for each pixel, signed bits at the times of sufficiently large brightness variations. 
Existing schemes for training WTA-SNNs are limited to rate-encoding solutions, and are hence able to detect only spatial patterns. 
Developing more general training algorithms for arbitrary WTA-SNNs inherits the challenges of training (binary) Spiking Neural Networks (SNNs). These amount, most notably, to the non-differentiability of threshold functions, to the recurrent behavior of spiking neural models, and to the difficulty of implementing backpropagation in neuromorphic hardware. 
In this paper, we develop a variational online local training rule for WTA-SNNs, referred to as {\VOWEL}, that leverages only local pre- and post-synaptic information for visible circuits, and an additional common reward signal for hidden circuits. 
The method is based on probabilistic generalized linear neural models, control variates, and variational regularization. 
Experimental results on real-world neuromorphic datasets with multi-valued events demonstrate the advantages of WTA-SNNs over conventional binary SNNs trained with state-of-the-art methods, especially in the presence of limited computing resources.
\end{abstract}

\begin{IEEEkeywords}
Probabilistic Spiking Neural Networks, Neuromorphic Computing
\end{IEEEkeywords}

\section{Introduction} \label{sec:intro}

From the pioneering works in the 40s and 50s, connectionist systems based on Artificial Neural Networks (ANNs) have made remarkable progress towards solving difficult tasks through supervision or feedback mechanisms. 
While reflecting the general distributed computational architecture of biological brains, ANNs approximate the operation of biological neurons by accounting for only one of the information-encoding mechanisms devised by brains \cite{borst1999information}. 
In fact, neurons in ANNs encode information in \emph{spatial} patterns of real-valued activations, which may be interpreted as modeling the spiking {\em rate} of biological neurons. 
Biological neurons, however, can encode and process information in \emph{spatio-temporal} patterns of temporally sparse spiking signals \cite{vanrullen2005spike}. 
With the recognition of the limitations of ANN-based solutions in terms of energy consumption \cite{hao2019training}, particularly for edge intelligence applications \cite{park2019wireless}, there is a renewed interest in exploring novel computational paradigms that can benefit from time encoding and temporally sparse, event-driven, processing \cite{intelspectrum}.

\begin{figure*}[ht!]
\centering
\subfigure[]{
\includegraphics[height=0.4\columnwidth]{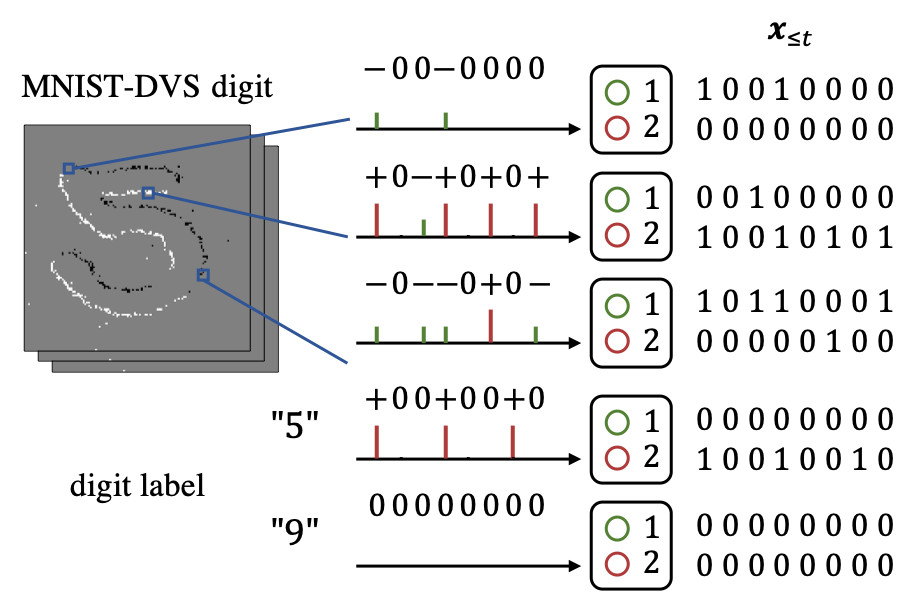}
\label{fig:wta-data}
}
\subfigure[]{
\includegraphics[height=0.4\columnwidth]{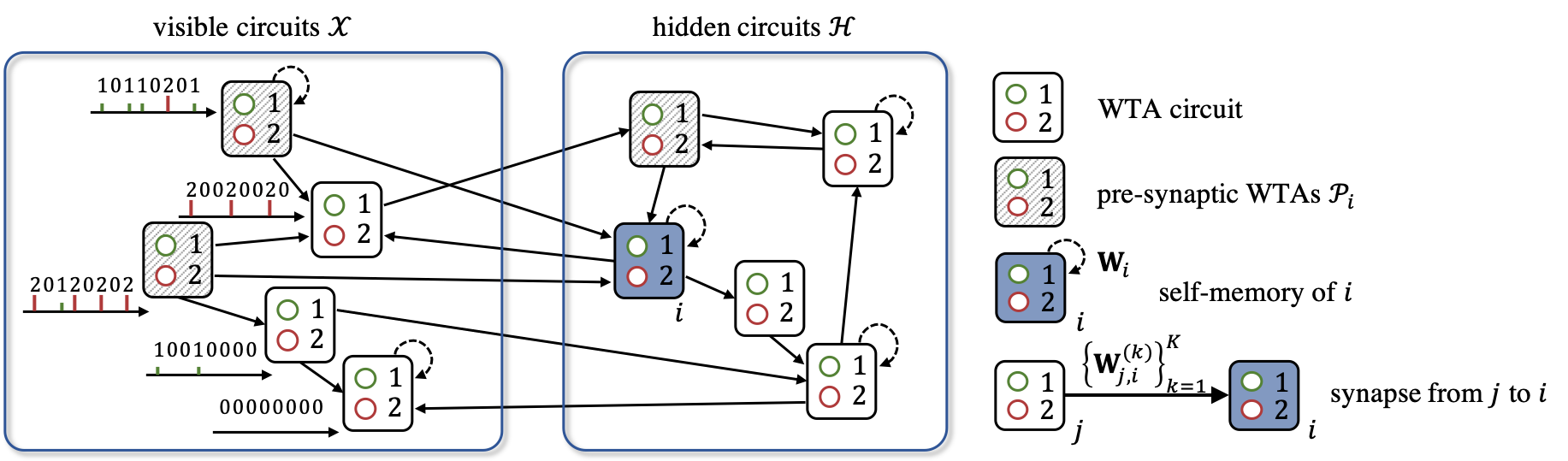}
\label{fig:wta-snn-architecture}
}
\vspace{-0.2cm}
\caption{Illustration of a WTA-SNN applied to a handwritten digit classification task using the MNIST-DVS dataset, with each WTA circuit $i$ having $C_i =2$ spiking units. (a) A moving MNIST digit captured with a neuromorphic camera records negative ($-$) or positive ($+$) events for each significant change in the pixel's luminosity. The negative event is encoded in the spiking of the first unit (green) of the corresponding circuit, while positive one is encoded by the second unit (red). (b) The directed links between two circuits represent synaptic dependencies, while the self-loop links represent self-memory. The directed graph may have loops, indicating recurrent behavior.}
\label{fig:wta-snn-example}
\vspace{-0.4cm}
\end{figure*}

{\bf Spiking Neural Networks (SNNs).} 
As a step towards capturing the temporal encoding capabilities of biological brains, Spiking Neural Networks (SNNs) have been introduced in the 90s \cite{maass1997networks}, inspiring a line of work that has crisscrossed the fields of neuroscience and machine learning \cite{churchland2016computational}. 
While initial research focused on aspects of biological plausibility, more recent activity has also targeted the definition of practical energy-efficient solutions for specific tasks, such as keyword spotting in audio signals \cite{blouw2019benchmarking} and object recognition in videos \cite{chadha2019neuromorphic}. 
These works have been largely motivated and supported by the significant progress made in the design of \emph{neuromorphic computing} platforms, such as IBM's TrueNorth, Intel's Loihi, and Brainchip's Akida. 

{\bf Challenges.} 
The design of training algorithms for SNNs faces a number of novel challenges as compared to the well-established state-of-the-art for ANNs. 
{\bf 1}) First, spiking neurons, if interpreted as deterministic devices as in ANNs, are characterized by \emph{non-differentiable} activation functions due to their threshold crossing-triggered behavior. 
{\bf 2}) Second, spiking neurons are internally \emph{recurrent} due to refractoriness mechanisms and recurrent topologies \cite{neftci2019surrogate}. 
Therefore, \emph{credit assignment} to hidden neurons -- whose behavior is not specified by data -- requires to propagate changes in the visible outputs through both space (neurons) and time. 
{\bf 3}) Third, most neuromorphic chips typically do not implement backpropagation paths, but only allow for \emph{local operations} along with global feedback signaling \cite{davies2018loihi}. 
{\bf 4}) Finally, groups of spiking neurons may spike in synchronized fashion, implementing forms of {\em population coding} through spatio-temporal correlations \cite{doya2007bayesian}. 
This paper aims at tackling all these challenges by leveraging a probabilistic framework grounded in variational learning.

{\bf State of the Art.} 
Solutions to the challenges outlined above vary. 
At one end, we have approaches that forego training on SNNs, which are then used solely as efficient inference machines. This is done by transferring weights from pre-trained ANNs \cite{rueckauer2018conversion} or from auxiliary non-spiking neural networks \cite{ponulak2010resume}. 
At the other end lie solutions that aim at the direct -- on chip, if implemented on hardware -- training of SNNs for the detection and reproduction of specific spatio-temporal patterns. 
Non-differentiability (challenge {\bf 1}) is tackled by smoothing out the activation function \cite{huh2018gradient} or its derivative \cite{neftci2019surrogate}, or by relying on probabilistic neural models \cite{jimenez2014stochastic, brea2013matching, jang19:spm}. 
Credit assignment (challenge {\bf 2}) may be carried out via backpropagation \cite{neftci2017eventbackprop, bohte2002error}; or, addressing also challenge {\bf 3}, via feedback alignment \cite{zenke2018superspike} or local randomized targets \cite{kaiser2018decolle}. 
For probabilistic models, local credit assignment is facilitated by variational inference \cite{jimenez2014stochastic, brea2013matching, jang19:spm}, as we further discuss in this paper. 
Finally, spatial correlation (challenge {\bf 4}) can be modeled via {\em Winner-Take-All} (WTA) circuits of spiking neurons, where each WTA circuit consists of a group of correlated spiking units, with at most one of the spiking units emitting a spike. To the best of our knowledge, as summarized in Section~\ref{sec:related}, existing work on WTA spiking circuits is currently limited to {\em rate-encoding} solutions \cite{guo2017hierarchical, mostafa2018learning}.



{\bf Contributions.} 
In this paper, we develop an online local training rule for arbitrary networks of spiking WTA circuits, referred to as {\bf V}ariational {\bf O}nline learning rule for spiking {\bf W}inner-tak{\bf E}-al{\bf L} circuits ({\bf{\VOWEL}}). 
The goal is to train the network to respond to input spatio-temporal patterns by producing desired sequences of spiking signals at its output. 
Consider, as an example, the set-up in Fig.~\ref{fig:wta-data}, where the input is provided by data produced by a neuromorphic camera. For each pixel, the camera produces a trace of signed event timings, with the timing marking the occurrence of a significant change in luminance and the sign describing the direction of the change. 
The network is trained to respond to images by producing specific spiking outputs that can be decoded into a class label. 
Importantly, encoding signed data requires the use of at least two correlated spiking neurons, motivating the use of WTAs. 
{\VOWEL} applies more generally to WTA circuits of any size.

The proposed training rule is derived via the maximization of a lower bound on the discounted log-likelihood function defined by assuming categorical generalized linear neural models \cite{jang19:spm, pillow2008spatio} for spiking WTAs. 
Unlike existing work, the approach applies to arbitrary topologies and spatio-temporal synaptic and somatic kernels (see Fig.~\ref{fig:wta-snn-architecture} and Fig.~\ref{fig:wta-membrane-potential}). 
{\VOWEL}, which generalizes the algorithms introduced in \cite{jimenez2014stochastic, brea2013matching, jang19:spm, zenke2018superspike} for conventional SNNs (without WTA circuits), follows the three-factor form \cite{fremaux2016neuromodulated}: A synapse weight $w_{j,i}$ from pre-synaptic neuron $j$ to a \emph{visible} post-synaptic neuron $i$ is updated as
\begin{subequations} \label{eq:biological-update}
\begin{align} \label{eq:biological-update-vis}
    w_{j,i} \leftarrow w_{j,i} + \eta \cdot \Big\langle  \text{post}_{i} \cdot \sum\nolimits_{k=1}^K \text{pre}_{j}^{(k)} \Big\rangle,
\end{align}
while a synaptic weight $w_{j,i}$ to a \emph{hidden} post-synaptic neuron $i$ is updated as
\begin{align} \label{eq:biological-update-hid}
    w_{j,i} \leftarrow w_{j,i} + \eta \cdot \Big\langle \ell \cdot \text{post}_{i} \cdot \sum\nolimits_{k=1}^K \text{pre}_{j}^{(k)} \Big\rangle,
\end{align}
\end{subequations}
where $\langle \cdot \rangle$ denotes a discounted time-averaging operator, and $\eta$ is a learning rate. The updates \eqref{eq:biological-update} depend on three types of factors. The first, $\text{post}_i$, is an error signal that depends on the activity of the post-synaptic neuron $i$. It is computed based on the desired behavior for a visible neuron $i$, and on model-driven sampled behavior for a hidden neuron $i$. The second term, $\sum_{k=1}^K \text{pre}_j^{(k)}$, amounts to the sum of $K$ distinct temporally filtered versions of the (desired or sampled) activity of the (visible or hidden) pre-synaptic neuron $j$. Finally, the scalar $\ell$ is a global reward signal that determines the sign and magnitude of the update for hidden neurons, as dictated by the current likelihood for the desired output of the visible neurons. 

Numerical experiments on standard datasets produced by a neuromorphic camera \cite{lichtsteiner2006128} demonstrate the capability of WTA-SNNs trained with {\VOWEL} to joint extract information from spatio-temporal patterns, and from numerical values assigned to each spike. Performance comparisons are presented with conventional SNNs trained with state-of-the-art algorithms \cite{kaiser2018decolle}. We specifically focus our evaluations on the resource-limited ``edge intelligence'' regime, characterized by small neural network topologies and large sampling periods for the acquisition of exogeneous signals \cite{davies2019spikemark}.

{\bf Notations.} We denote negative cross-entropy of two non-negative vectors ${\bm a}$ and ${\bm b}$ with $\textstyle\sum\nolimits_x a_x \leq 1$ and $\textstyle\sum\nolimits_x b_x \leq 1$ by   
\begin{align*} 
    \bar{H}({\bm a}, {\bm b}) := \textstyle \sum\limits_x a_x \log b_x + (1-\sum\limits_x a_x) \log \Big(1 - \sum\limits_x b_x\Big),
\end{align*}
and the Kullback-Liebler (KL) divergence as $\text{KL}({\bm a} || {\bm b}) = \bar{H}\big({\bm a},{\bm a}\big) - \bar{H}\big({\bm a}, {\bm b}\big)$. The temporal average $\langle f_t \rangle_\kappa$ of a time sequence $\{f_t\}_{t \geq 1}$ with constant $\kappa \in (0,1)$ is defined as $\langle f_t \rangle_\kappa = \kappa \cdot \langle f_{t-1} \rangle_\kappa + f_t,$ with $\langle f_0\rangle_\kappa = 0$. 
Finally, we denote by $f_t \ast g_t = \sum_{\delta > 0} f_\delta g_{t-\delta}$ the convolution operator. 

\section{Background} \label{sec:background}

In this section, we introduce the operational principles of SNNs and the key definitions, and review the online local variational training rules derived in \cite{jimenez2014stochastic, brea2013matching, jang19:spm} for conventional SNNs (not including WTA circuits).

{\bf SNNs as Recurrent Binary Neural Networks.} 
An SNN is defined by a network of spiking neurons connected over an arbitrary graph, possibly including (directed) cycles. 
Focusing on a discrete-time implementation, as in most hardware solutions \cite{davies2018loihi}, each spiking neuron $i$ at discrete time $t=1,2,...$ outputs a binary value $s_{i,t} \in \{0,1\}$, with ``1'' denoting the firing of a spike. We collect in vector $\bms_t = (s_{i,t}: i \in \set{V})$ the spikes emitted by all neurons $\set{V}$ at time $t$ and denote by $\bms_{\leq t} = (\bms_1, \ldots, \bms_t)$ the spike sequences of all neurons up to time $t$. Each neuron $i$ receives inputs from the set $\set{P}_i$ of pre-synaptic neurons, which are connected to it via directed links in the graph. The current output of the neuron depends on a function 
\begin{align} \label{eq:binary-output}
s_{i,t} = f_{\bmtheta_i}(\bms_{\set{P}_i, \leq t-1}, \bms_{i,\leq t-1}, z_{i,t})
\end{align} 
of the sequences $\bms_{\set{P}_i, \leq t-1}$ emitted by pre-synaptic neurons $\set{P}_i$, of the local spiking history $\bms_{i,\leq t-1}$, and, possibly, of a source of randomness $z_{i,t}$. The function $f_{\bmtheta_i}(\cdot)$, which defines the recurrent operation of a neuron, is generally non-differentiable in the local model parameters $\bmtheta_i$, since it describes a threshold activation that dictates whether the neuron spikes ($s_{i,t}=1$) or is silent ($s_{i,t}=0$). 
While expression \eqref{eq:binary-output} allows for a general dependence on the spiking history, in practice, function $f_{\bmtheta_i}(\cdot)$ depends on a number of state variables, including synaptic and somatic traces, that summarize the past of spiking signals $\bms_{\set{P}_i, \leq t-1}$ and $\bms_{i, \leq t-1}$ \cite{neftci2019surrogate}. Deterministic models disable the dependence on randomness by setting $z_{i,t} = 0$. As mentioned above, simplifications \cite{neftci2019surrogate, huh2018gradient, neftci2017eventbackprop, zenke2018superspike, kaiser2018decolle} for deterministic models yield local update rules, which we relate to the proposed approach in Sec.~\ref{sec:vem-wta}.

{\bf Generalized Linear Model (GLM).} 
As an alternative to deterministic models, probabilistic GLMs enable the principled derivation of online local learning rules via the maximization of (lower bounds on) the likelihood function. By leveraging the source of randomness $z_{i,t}$, GLMs define the spiking probability of neuron $i$ at time $t$ as 
\begin{align} \label{eq:binary-ind-prob}
    p_{\bmtheta_i}(s_{i,t} = 1 | \bms_{\leq t-1}) = p_{\bmtheta_i}(s_{i,t}=1|u_{i,t}) = \sigma(u_{i,t}),
\end{align}
with $\sigma(x)=(1+e^{-x})^{-1}$ being the sigmoid function and the membrane potential $u_{i,t}$ being a (deterministic) function of the state variables dependent on the past samples $\bms_{\set{P}_i, \leq t-1}$ and $\bms_{i, \leq t-1}$. Note that the spiking probability \eqref{eq:binary-ind-prob} increases with the membrane potential. 
From \eqref{eq:binary-ind-prob}, the log-probability corresponds to the binary negative cross-entropy, i.e., 
\begin{align} \label{eq:binary-ind-ce}
    &\log p_{\bmtheta_i}(s_{i,t} | u_{i,t}) = \bar{H} \big(s_{i,t}, \sigma(u_{i,t}) \big).
\end{align}
The joint probability of the spike signals $\bms_{\leq T}$ emitted by all neurons up to time $T$ is defined using the chain rule as $p_{\bmtheta}(\bms_{\leq T}) = \prod_{t=1}^T \prod_{i \in \set{V}} p_{\bmtheta_i}(s_{i,t} | u_{i,t})$, 
where $\bmtheta = \{\bmtheta_i\}_{i \in \set{V}}$ is the model parameters.
        
Various models can be considered for the membrane potential $u_{i,t}$ that entail distinct memory requirements. Here we follow the approach in, e.g., \cite{doya2007bayesian}, and assume that the membrane potential is obtained as the output of spatio-temporal moving average filters with finite-duration for both synapses and self-memory. Specifically, to account for synaptic memory, we pre-define, as part of the inductive bias, $K$ finite-duration filters $\{a_t^{(k)}\}_{k=1}^K$, and, for neural self-memory, we similarly introduce a finite-duration filter $b_t$ (multiple somatic filters can also be considered). Each $(j,i)$ synapse between pre-synaptic neuron $j$ and post-synaptic neuron $i$ computes the synaptic filtered trace $\overra{s}_{j,t}^{(k)} = a_t^{(k)} \ast s_{j,t}$, 
while the soma of each neuron $i$ computes the feedback, or self-memory, trace $\overla{s}_{i,t} = b_t \ast s_{i,t}$.
The membrane potential of neuron $i$ at time $t$ is then given as the weighted sum
\begin{equation} \label{eq:binary-potential}
    u_{i,t} = \sum_{j \in \set{P}_i} \sum_{k=1}^K w_{j,i}^{(k)} \overra{s}_{j,t-1}^{(k)} + w_i \overla{s}_{i,t-1} + \vartheta_i,
\end{equation}
where $\{w_{j,i}^{(k)}\}_{k=1}^K$ is the set of learnable synaptic weights from pre-synaptic neuron $j \in \set{P}_i$ to post-synaptic neuron $i$; $w_i$ is the learnable feedback weight; and $\vartheta_i$ is a learnable bias parameter, with $\bmtheta_{i} = \{ \{\{w_{j,i}^{(k)}\}_{k=1}^K\}_{j \in \set{P}_i}, w_i, \vartheta_i \}$ being the local model parameters. We observe that the GLM-based SNN outlined above can be interpreted as a dynamic version of belief networks \cite{neal1992beliefnets} and can also be interpreted as an autoregressive discrete process.

{\bf Training GLM-based SNNs.} 
As is common in probabilistic models \cite{koller2009probabilistic}, learning can be carried out by maximizing the likelihood (ML) that a subset $\set{X}$ of ``visible'' neurons outputs desired spiking signals $\bmx_{\leq T}$ in response to given inputs. Mathematically, we can write the problem as $\max_{\bmtheta} \log p_{\bmtheta}(\bmx_{\leq T})$. Importantly, the stochastic spiking signals $\bmh_{\leq T}$ of ``hidden'' neurons in the complementary set $\set{H}$ have to be averaged over to evaluate the likelihood $\log p_{\bmtheta}(\bmx_{\leq T}) = \log \sum_{\bmh_{\leq T}} p_{\bmtheta}(\bmx_{\leq T}, \bmh_{\leq T})$. To address the problem, {\em variational Expectation-Maximization (VEM)} maximizes the lower bound  
\begin{align} \label{eq:binary-elbo}
    &L_{\bmx_{\leq T}}(\bmtheta) := \mathbb{E}_{q(\bmh_{\leq T})} \Big[ \log \frac{p_{\bmtheta}(\bmx_{\leq T}, \bmh_{\leq T})}{q(\bmh_{\leq T})} \Big] 
\end{align}
over both the model parameters $\bmtheta$ and the variational posterior distribution $q(\bmh_{\leq T}) = \prod_{t=1}^T q(\bmh_t | \bmh_{\leq t-1})$. 
        
As proposed in \cite{jimenez2014stochastic, brea2013matching, jang19:spm}, we choose the variational distribution to equal the ``causally conditioned'' distribution of the hidden neurons given the visible neurons\footnote{The notion of causally conditioned distribution was introduced in the information-theoretic literature in \cite{kramer1998directed}.}
\begin{eqnarray} \label{eq:binary-caually-cond}
    q(\bmh_{\leq T}) = p_{\bmtheta^\text{H}}(\bmh_{\leq T} || \bmx_{\leq T-1}) = \prod_{t=1}^T \prod_{i \in \set{H}} p_{\bmtheta_i}(h_{i,t}|u_{i,t}).
\end{eqnarray} 
Note that we indicate as $\bmtheta^\text{X} = \{\bmtheta_i\}_{i \in \set{X}}$ and $\bmtheta^\text{H} = \{\bmtheta_i\}_{i \in \set{H}}$ the collection of model parameters for visible and hidden neurons respectively. In contrast to true posterior $p_{\bmtheta}(\bmh_{\leq T}|\bmx_{\leq T-1})$, distribution \eqref{eq:binary-caually-cond} ignores the dependence of the hidden neurons' outputs on future samples of the visible neurons. With this choice, the ratio in \eqref{eq:binary-elbo} equals to the binary negative cross-entropy loss for the visible neurons, i.e., 
\begin{align} \label{eq:binary-ls}
    \log \frac{p_{\bmtheta}(\bmx_{\leq T}, \bmh_{\leq T})}{q(\bmh_{\leq T})} 
    &= \sum_{t=1}^T \sum_{i \in \set{X}} \bar{H}\big(x_{i,t}, \sigma(u_{i,t})\big).
\end{align}

Furthermore, in order to obtain online learning rules, we specifically aim at maximizing at each time $t$ a discounted version of the lower bound \eqref{eq:binary-elbo} with \eqref{eq:binary-caually-cond}. Online local update rules for the model parameter can be obtained by considering the maximization of this criterion via stochastic gradient descent \cite{jimenez2014stochastic, brea2013matching}, \cite[Algorithm~$2$]{jang19:spm}, which we will generalize for networks of spiking WTA circuits in Sec.~\ref{sec:vem-wta}. For completeness, we provide details for these online learning rules in the Supplementary Material.  

\section{{\VOWEL}: Online Local VEM-WTA Rule} \label{sec:vem-wta}

In this section, we derive and detail the online local learning rule for arbitrary networks of spiking WTA circuits based on variational learning. 

\subsection{Probabilistic WTA-SNNs}

{\bf Recurrent Networks of Spiking WTAs.} A WTA-SNN is an arbitrary directed network of spiking WTA circuits. As for SNNs, the connectivity graph can possibly include (directed) cycles. Each WTA circuit $i$ consists of $C_i$ correlated spiking units, with {\em at most} one of the $C_i$ spiking units emitting a spike, i.e., a ``1'', at any discrete time $t$. Note that in general each WTA circuit may include a different number of units, and that a WTA circuit $i$ with $C_i=1$ is a conventional spiking neuron. Mathematically, we define the output of WTA $i$ at time $t$ by the $C_i \times 1$ vector
\begin{equation} \label{eq:wta-onehot}
    \bms_{i,t} = 
    \begin{cases}
    \mathbf{0} ~~~ \text{if WTA $i$ does not fire at time $t$},\\
    \bfe_c ~~ \text{if the $c$th unit spikes at time $t$},
    \end{cases}
\end{equation}
where $\bfe_c$ is a $C_i \times 1$ vector that has all-zero entries except for the $c$th entry equal to $1$. As in Sec.~\ref{sec:background}, we collect in vector $\bms_t = (\bms_{i,t}: i \in \set{V})$ the signals emitted by all WTA circuits $\set{V}$ at time $t$, and denote $\bms_{\leq T} = (\bms_1, \ldots, \bms_T)$ all the signals in the interval $t \in \{1, \ldots, T\}$ for some $T > 0$.

{\bf Probabilistic GLM for WTA-SNNs.} 
In a manner similar to \eqref{eq:binary-output}, for each WTA circuit $i$, one can define a general transfer function $f_{\bmtheta_i}(\cdot)$ mapping past behaviors $\{ \bms_{\set{P}_i, \leq t-1}, \bms_{i, \leq t-1}\}$ of pre-synaptic WTA circuits $\set{P}_i$ and of the circuit $i$ to the current vector output \eqref{eq:wta-onehot}. A deterministic model for WTA circuits would require the use of a multi-valued quantizer or smooth approximation thereof. 
In this paper, we instead adopt a probabilistic framework that allows us to directly derive local online rules in a principled fashion. 

\begin{figure}[t!]
\begin{center}
\centerline{\includegraphics[width=1\columnwidth]{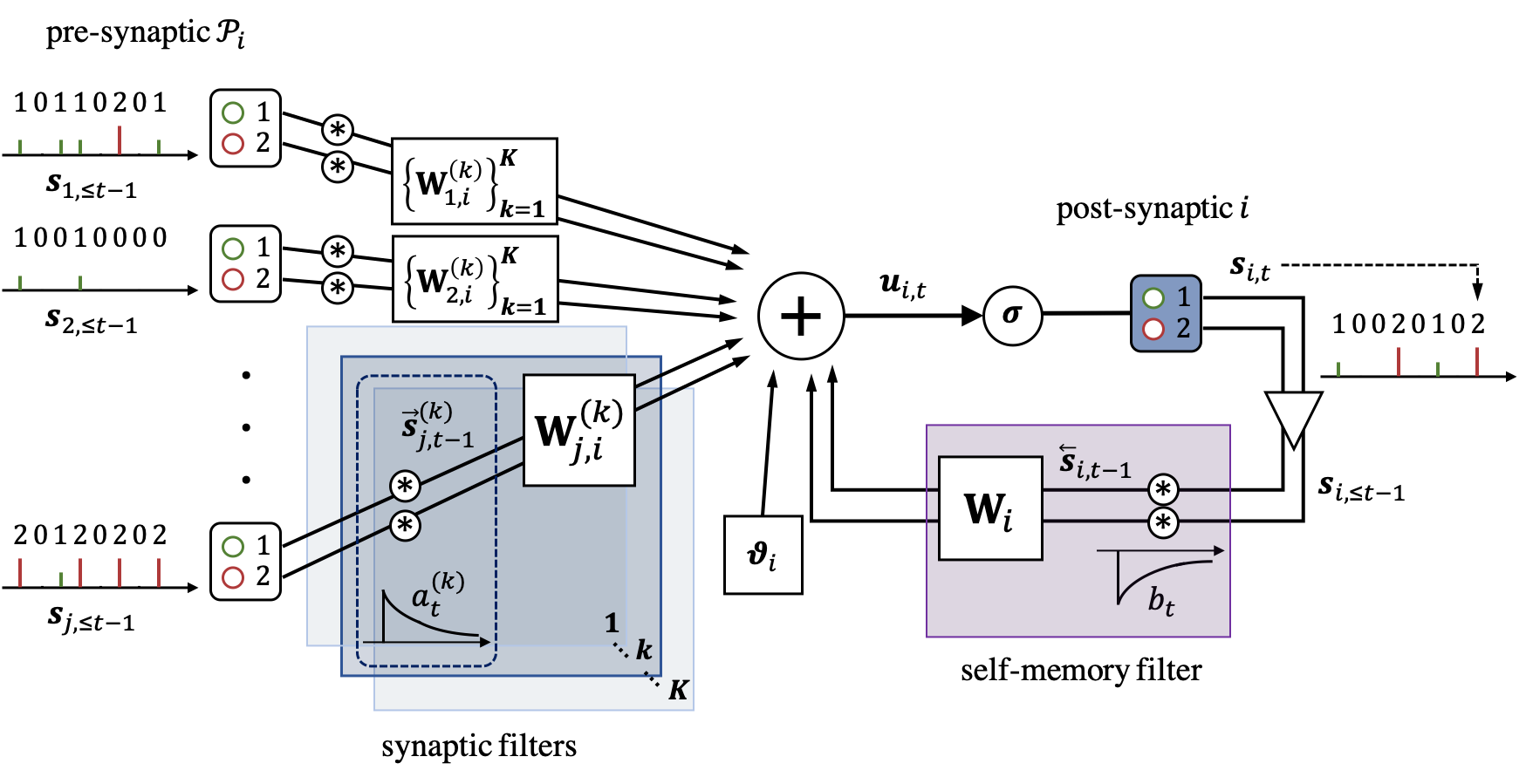}}
\vspace{-0.3cm}
\caption{An illustration of the membrane potential $\bmu_{i,t}$ model for a WTA-SNN, with $C_i = 2$ units for all circuits and exponential synaptic and somatic filters. The contributions of the synaptic traces from a pre-synaptic WTA circuit $j \in \set{P}_i$ through $K$ synaptic filters $\{a_t^{(k)}\}_{k=1}^K$ are multiplied by the corresponding weight matrices $\{\bfW_{j,i}^{(k)}\}_{k=1}^K$, and then summed over $k=1,\ldots,K$. The contribution of the somatic trace of a post-synaptic circuit $i$ through somatic filter $b_t$ is multiplied by a weight matrix $\bfW_i$. The bias parameter $\bmvartheta_i$ is summed to obtain the membrane potential $\bmu_{i,t}$, which is used to determine the distribution of the WTA circuit output $\bms_{i,t}$ through the $\bmsigma(\cdot)$ operator.
}
\label{fig:wta-membrane-potential}
\end{center}
\vspace{-0.7cm}
\end{figure}

A probabilistic GLM for WTA circuits defines the probability of circuit $i$ to output a spike at unit $c$ for time $t$ as
\begin{align*} 
    p_{\bmtheta_i}(\bms_{i,t} = \bfe_c | \bmu_{i,t}) = \sigma_c(\bmu_{i,t}) := \frac{ \exp\big( \bmu_{i,t}^\top \bfe_c \big) }{ 1 + \sum_{c'=1}^C \exp\big( \bmu_{i,t}^\top \bfe_{c'} \big) },
\end{align*}
where the dependence on history $\{ \bms_{\set{P}_i, \leq t-1}, \bms_{i, \leq t-1} \}$ is mediated by the $C_i \times 1$ membrane potential vector $\bmu_{i,t}$, and we define the vector function $\bmsigma(\bmu_{i,t}) = \big[ \sigma_1(\bmu_{i,t}), \ldots, \sigma_C(\bmu_{i,t}) \big]^\top$. Note that the spiking probability for unit $c$ in circuit $i$ increases with the $c$th entry value of the membrane potential $\bmu_{i,t}$. Under this model, the log-probability equals the categorical negative cross-entropy
\begin{equation} \label{eq:wta-ind-ce}
\log p_{\bmtheta_i}(\bms_{i,t}|\bmu_{i,t}) = \bar{H} \big( \bms_{i,t}, \bmsigma(\bmu_{i,t}) \big). 
\end{equation}

Following a natural generalization of the membrane potential model for binary SNNs in \eqref{eq:binary-potential}, as illustrated in Fig.~\ref{fig:wta-membrane-potential}, we assume pre-defined $K$ finite-duration spatio-temporal filters $\{a_t^{(k)}\}_{k=1}^K$ and $b_t$ for synapses and self-memory, respectively. Accordingly, the synaptic traces $\{\overra{\bms}_{j,t}^{(k)}\}_{k=1}^K$ for each synapse $(j,i)$ from pre-synaptic circuit $j$ to post-synaptic circuit $i$ are defined by the $C_j \times 1$ vectors
\begin{equation*}
    \overra{\bms}_{j,t}^{(k)} = a_t^{(k)} \ast \bms_{j,t}, 
\end{equation*}
and the somatic trace is similarly given by the $C_i \times 1$ vector $\overla{\bms}_{i,t} = b_t \ast \bms_{i,t}$. The membrane potential of circuit $i$ at time $t$ is a $C_i \times 1$ vector given as the weighted sum 
\begin{eqnarray} \label{eq:wta-potential}
    \bmu_{i,t} = \sum_{j \in \set{P}_i} \sum_{k=1}^K \bfW_{j,i}^{(k)} \overra{\bms}_{j,t-1}^{(k)} + \bfW_i \overla{\bms}_{i,t-1} + \bmvartheta_i,
\end{eqnarray}
where $\{\bfW_{j,i}^{(k)}\}_{k=1}^K$ is the set of learnable synaptic weight matrices of size $C_i \times C_j$ for synapse $(j,i)$; $\bfW_i$ is the learnable feedback weight matrix of size $C_i \times C_i$; and $\bmvartheta_i$ is a learnable $C_i \times 1$ bias parameter, with $\bmtheta_i = \{ \{\{\bfW_{j,i}^{(k)}\}_{k=1}^K\}_{j \in \set{P}_i}, \bfW_i, \bmvartheta_i \}$ being the local model parameters of circuit $i$. Note that each synapse is here defined by a matrix weight for each filter, and not by a scalar as in the binary case. 

A WTA circuit can be implemented as a group of spiking neurons connected through an inhibition mechanism that ensures the spiking of a single neuron \cite{guo2017hierarchical, oster2006spiking, nessler2013bayesian}. According to the model \eqref{eq:wta-potential}, synapses connect circuits and are generally capable of matrix-vector multiplication. Note that, due to the binary nature of vectors $\{\bms_{i,t}\}$, no multiplications are required during inference, i.e., to evaluate \eqref{eq:wta-potential}. As for conventional SNNs, the energy consumption during inference hence depends on the number of spikes, i.e., non-zero vectors, communicated across the synapses between circuits.  

\subsection{{\VOWEL}}

{\bf Training GLM-based WTA-SNN.} 
For training, we partition the set $\set{V}$ of WTA circuits into disjoint subsets $\set{X}$ and $\set{H}$ of visible and hidden circuits, respectively. By following analogous steps as \eqref{eq:binary-elbo}-\eqref{eq:binary-ls} for binary SNNs, we obtain the following lower bound on the log-likelihood of the desired behavior $\bmx_{\leq T}$ of the visible circuits 
\begin{equation*} 
    L_{\bmx_{\leq T}}(\bmtheta) := \mathbb{E}_{p_{\bmtheta^\text{H}}(\bmh_{\leq T}||\bmx_{\leq T-1})}\bigg[ \sum_{t=1}^{T} \sum_{i \in \set{X}} \bar{H}\big( \bmx_{i,t}, \bmsigma(\bmu_{i,t}) \big) \bigg].
\end{equation*}

{\bf Sparsity-inducing Variational Regularization.} 
While the behavior of visible circuits is specified by training example $\bmx_{\leq T}$, the behavior $\bmh_{\leq T}$ of hidden circuits in a WTA-SNN may present dense spiking signals since the ML criterion $L_{\bmx_{\leq T}}(\bmtheta)$  
does not enforce any sparsity constraint. 
To obviate this problem, bounded rationality \cite{leibfried2015reward} regularization (or generalized Bayesian inference \cite{knoblauch2019generalized}) can be applied to the ML problem by adding a term that penalizes deviations of the hidden circuits' distribution $p_{\bmtheta^\text{H}}(\bmh_{\leq T}||\bmx_{\leq T-1})$ from a baseline distribution $r(\bmh_{\leq T})$ with the desired level of sparsity $r$. 
Mathematically, the learning criterion is given as 
\begin{align} \label{eq:wta-elbo-reg-online}
    &L_{\bmx_{\leq t}}(\bmtheta) - \alpha \cdot \text{KL}\big( p_{\bmtheta^\text{H}}(\bmh_{\leq t} || \bmx_{\leq t-1}) || r(\bmh_{\leq t}) \big) \cr 
    &~~ = \mathbb{E}_{p_{\bmtheta^\text{H}}(\bmh_{\leq t}||\bmx_{\leq t-1})} \bigg[ \sum_{t'=0}^{t-1} \gamma^{t'} \ell_{\bmtheta, t-t'} \bigg] := L_{\bmx_{\leq t}}^\text{reg}(\bmtheta),
\end{align}
where $\alpha > 0$ is a regularization coefficient; we have defined the learning reward $\ell_{\bmtheta,t}$ at time $t$ as \begin{align} \label{eq:wta-ls}
    \ell_{\bmtheta,t} &= \sum_{i \in \set{X}} \bar{H}\big(\bmx_{i,t}, \bmsigma(\bmu_{i,t}) \big) \cr 
    &~~ - \alpha \sum_{i \in \set{H}} \Big( \bar{H}\big( \bmh_{i,t}, \bmsigma(\bmu_{i,t})\big) - \log r(\bmh_{i,t}) \Big);
\end{align}
and we have introduced a weighted time average with discount factor $\gamma \in (0,1)$ for online learning.

{\bf Variational Online learning rule for spiking Winner-takE-AlL circuits ({\VOWEL}).} 
{\VOWEL} maximizes the lower bound $L_{\bmx_{\leq t}}^\text{reg}(\bmtheta)$ in \eqref{eq:wta-elbo-reg-online} via stochastic gradient descent as $\bmtheta \leftarrow \bmtheta + \eta \cdot \grad_{\bmtheta} \hat{L}_{\bmx_{\leq t}}^\text{reg}(\bmtheta)$, in the direction of a Monte Carlo (MC), stochastic, estimate $\grad_{\bmtheta} \hat{L}_{\bmx_{\leq t}}^\text{reg}(\bmtheta)$. This requires the gradient of the log-probability \eqref{eq:wta-ind-ce} for each circuit $i$ with respect to its local parameters $\bmtheta_i$, which is given as 
\begin{subequations} \label{eq:wta-ll-grad}
\begin{align}
    &\grad_{\bmvartheta_i} \log p_{\bmtheta_i}(\bms_{i,t}|\bmu_{i,t}) = \underbrace{\bms_{i,t} - \bmsigma(\bmu_{i,t})}_{\text{post}_i}, \label{eq:wta-ll-grad-bias} \\
    &\grad_{\bfW_{j,i}^{(k)}} \log p_{\bmtheta_i}(\bms_{i,t}|\bmu_{i,t}) = \underbrace{\big( \bms_{i,t} - \bmsigma(\bmu_{i,t}) \big)}_{\text{post}_i} \underbrace{ \big(\overra{\bms}_{j,t-1}^{(k)}\big)^\top}_{\text{pre}_j^{(k)}}, \label{eq:wta-ll-grad-synaptic} \\
    &\grad_{\bfW_i} \log p_{\bmtheta_i}(\bms_{i,t}|\bmu_{i,t}) = \underbrace{\big( \bms_{i,t} - \bmsigma(\bmu_{i,t}) \big) \overla{\bms}_{i,t-1}^\top}_{\text{post}_i}. \label{eq:wta-ll-grad-feedback}
\end{align}
\end{subequations}
In order to estimate the gradient of learning criterion \eqref{eq:wta-elbo-reg-online}, at each time $t$, each hidden circuit $i \in \set{H}$ generates a sample $\bmh_{i,t} \sim p_{\bmtheta_i}(\bmh_{i,t}|\bmu_{i,t})$ from the causally conditioned distribution \eqref{eq:binary-caually-cond}. Then, for visible circuits $i \in \set{X}$, {\VOWEL} updates the model parameters by an amount equal to 
\begin{subequations} \label{eq:wta-online-update}
\begin{eqnarray} \label{eq:wta-online-update-vis}
    \Delta \bmtheta_i = \eta \cdot \Big\langle \underbrace{\grad_{\bmtheta_i} \log p_{\bmtheta_i}(\bmx_{i,t}|\bmu_{i,t})}_{\text{pre and post}} \Big\rangle_{\gamma},
\end{eqnarray}
while, for hidden circuits $i \in \set{H}$, we have the update
\begin{eqnarray} \label{eq:wta-online-update-hid}
    \Delta \bmtheta_i = \eta \cdot \Big\langle \underbrace{\big( \ell_{\bmtheta,t} - \bmb_{i,t} \big)}_{\text{reward}} \cdot ~\big\langle \underbrace{\grad_{\bmtheta_i} \log p_{\bmtheta_i}(\bmh_{i,t}|\bmu_{i,t})}_{\text{pre and post}} \big\rangle_{\kappa} \Big\rangle_{\gamma},
\end{eqnarray}
\end{subequations}
with some constant $\kappa \in (0,1)$. The baseline $\bmb_{i,t}$ is a control variate introduced as a means to reduce the variance of estimate. We will discuss this further at the end of this section. Details of the proposed learning algorithm with derivation of the gradient estimates are provided in the Supplementary Material.

\subsection{Interpreting {\VOWEL}}

Following the discussion of \eqref{eq:biological-update} in Sec.~\ref{sec:intro}, the update rule \eqref{eq:wta-online-update} is local, with the exception of the global scalar reward signal $\ell_{\bmtheta,t}$ used for the update of hidden circuits' parameters. 

{\bf Pre-synaptic trace.} In the update of each synaptic weight matrix $\bfW_{j,i}^{(k)}$, the pre-synaptic circuit $j$ contributes a term equal to the synaptic filtered trace $\overra{\bms}_{j,t-1}^{(j)}$. A lack of recent spiking activity from circuit $j$ hence yields a vanishing synaptic update. Specifically, if a unit $c$ of pre-synaptic circuit $j$ has not spiked within the memory of the synaptic kernels of the temporal averages in \eqref{eq:wta-online-update}, then, by \eqref{eq:wta-ll-grad-synaptic}, the update of the $c$th column of matrix $\bfW_{j,i}^{(k)}$ is approximately zero. This pre-synaptic term is also common to rules derived from deterministic SNN models (i.e., with $C_i = 1$ for all $i \in \set{V}$) such as \cite{huh2018gradient, bohte2002error, zenke2018superspike}.

{\bf Post-synaptic error.} 
The updates of model parameters for each circuit $i$ in \eqref{eq:wta-ll-grad} depend on the post-synaptic error term $\bms_{i,t} - \bmsigma(\bmu_{i,t})$, which plays a different role for visible and hidden circuits. For visible circuits, it provides a feedback signal in terms of the gap between the desired output $\bmx_{i,t}$, to which visible circuits are clamped during learning, and the average behavior $\bmsigma(\bmu_{i,t})$ under the current model. For hidden circuits, this term amounts instead to the difference between the current, randomly generated, output $\bmh_{i,t}$ and the average model behavior. This term hence magnifies updates for the time instants in which desired or current behavior differs significantly for the average counterpart -- an instance of {\em homeostatic plasticity} \cite{habenschuss2012homeostatic}. 
We note that, in rules derived from deterministic binary models, the post-synaptic term is given solely by the actual output neural value, possibly multiplied by an approximate activation derivative \cite{zenke2018superspike, kaiser2018decolle}. As such, it does not provide any form of learning signal.

{\bf Global reward signal.} 
For hidden circuits, the common global reward signal $\ell_{\bmtheta,t}$ in \eqref{eq:wta-ls} is used to guide the update \eqref{eq:wta-online-update-hid}. 
The global reward term \eqref{eq:wta-ls} indicates how effective the current, randomly sampled, behavior of hidden circuits is in ensuring the maximization of the likelihood of the desired behavior for the visible circuits. The global reward can be computed by a central processor by collecting the membrane potentials $\bmu_{i,t}$ from all circuits. In rules derived from deterministic binary models, a global error signal is instead provided to {\em all} neurons by following various heuristics, such as feedback alignment \cite{zenke2018superspike} or local per-layer rewards \cite{kaiser2018decolle, mostafa2018learning}.

{\bf Optimized baseline for variance reduction.} 
In order to reduce the variance of the gradient estimate \eqref{eq:wta-online-update-hid} for hidden circuits, we adopt an optimized baseline that minimizes a bound on the variance of the gradient \cite{peters2008reinforcement}. The optimized baseline can be obtained as $\bmb_{i,t} = \langle \ell_{\bmtheta,t} \cdot \bme_{i,t}^2 \rangle / \langle \bme_{i,t}^2 \rangle$, 
where we denoted $\bme_{i,t} = \langle \grad_{\bmtheta_i} \log p_{\bmtheta_i}(\bmh_{i,t}|\bmu_{i,t})\rangle_\kappa$. We note that the baseline is defined and updated locally at each circuit and separately for each local parameter.

\section{Experiments} \label{sec:exp}

\subsection{Experimental Setting}

We evaluate {\VOWEL} on two classification tasks defined by the neuromorphic datasets MNIST-DVS \cite{serrano2015poker} and DVSGesture \cite{davies2019spikemark, amir2017dvsgesture}. Both datasets have signed spiking signals as inputs and class labels as outputs (see Fig.~\ref{fig:wta-data}). The spiking sequences are obtained by sampling the input signals as detailed in the Supplementary Material. 
We consider a generic, non-optimized, network architecture characterized by a set of $H$ fully connected hidden circuits, all receiving the exogeneous inputs as pre-synaptic signals, and a read-out layer. Each circuit in the read-out layer receives pre-synaptic signals from all the hidden circuits, as well as from the exogeneous inputs. All WTA circuits in the WTA-SNN have the same number $C=2$ of spiking units to represent the signed spiking signal.

As benchmarks, we consider standard binary SNNs trained with {\VOWEL} in the special case $C=1$, as well as with the recently proposed {\DECOLLE} algorithm \cite{kaiser2018decolle}, which is derived based on a deterministic SNN model. For binary SNNs, the inputs' signs can be either discarded, producing a binary input signal (i.e., spike or silent) per pixel (as in, e.g., \cite{zhao2014feedforward}); or else two binary signals per pixel can be treated as separate binary inputs (as in, e.g., \cite{henderson2015spike}). We refer to the first case as ``unsigned binary inputs'' and to the second as ``per-sign binary inputs''. The mentioned benchmark binary SNNs are trained with $2H$ hidden neurons for a comparison with equal number of spiking units. Other hyperparameters and training settings are detailed in the Supplementary Material.  


\subsection{MNIST-DVS Classification} \label{subsec:MNISTDVS}

First, we consider a handwritten digit classification task based on the MNIST-DVS dataset \cite{serrano2015poker}. For this task, we train WTA-SNNs with $10$ spiking circuits in the read-out layer, one for each label. 

\begin{figure}[t!]
\begin{center}
\centerline{\includegraphics[width=.92\columnwidth]{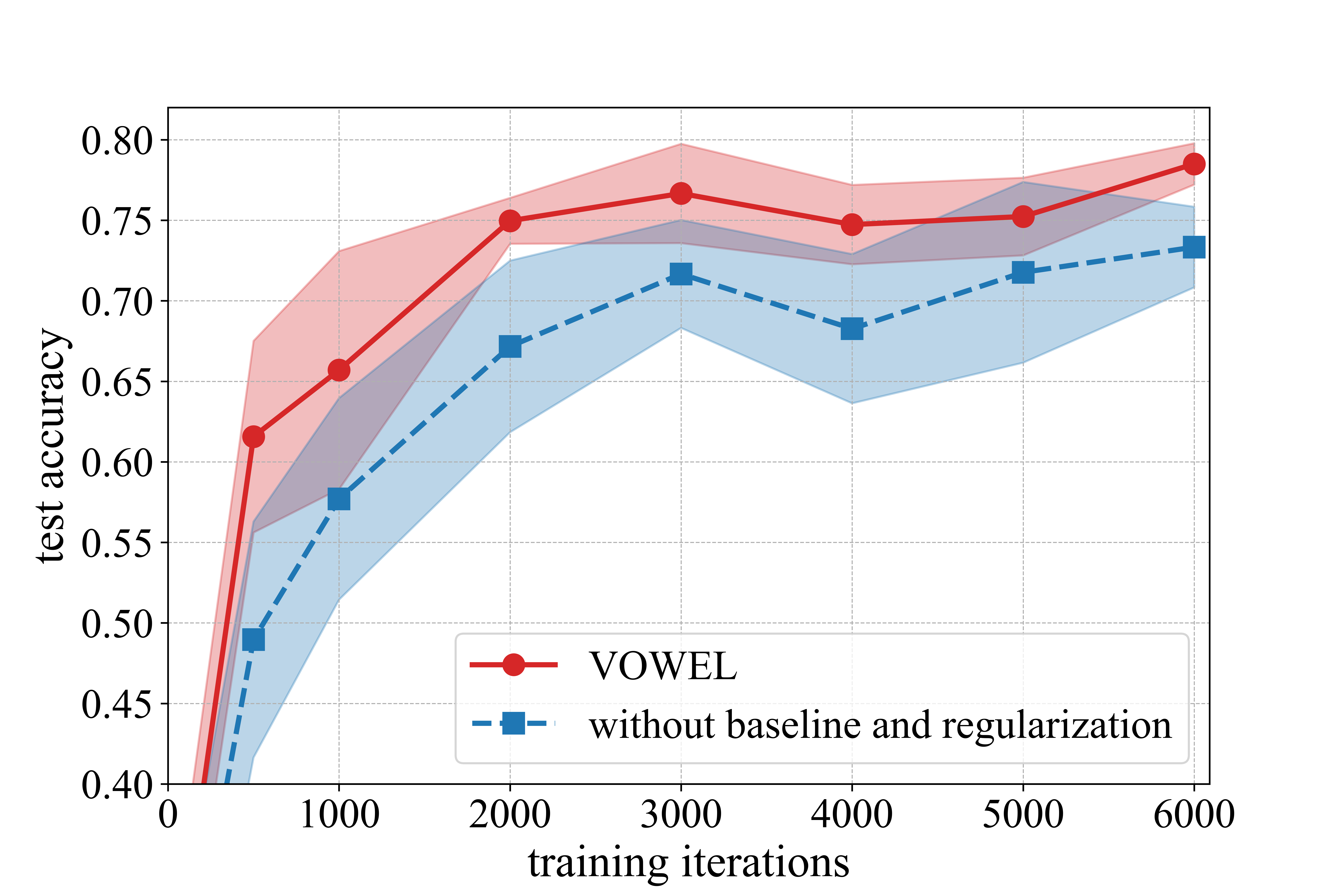}}
\vspace{-0.3cm}
\caption{Test accuracy for {\VOWEL} with $H=16$ hidden circuits for the MNIST-DVS dataset with and without baseline and regularization. Shaded areas represent confidence intervals equal to one standard deviation.}
\label{fig:baesline_reg}
\end{center}
\vspace{-0.6cm}
\end{figure}


To start, Fig.~\ref{fig:baesline_reg} compares the test accuracy performance of {\VOWEL} with a modified version of {\VOWEL} without regularization ($\alpha = 0$) and baseline $(\bmb_{i,t} = {\bm 0})$. It can be observed that baseline and regularization can improve the test performance both in terms of average accuracy, by reducing overfitting, and of robustness, by reducing the variance of the stochastic gradients \eqref{eq:wta-online-update}.

\begin{figure}[t]
\begin{center}
\centerline{\includegraphics[width=.92\columnwidth]{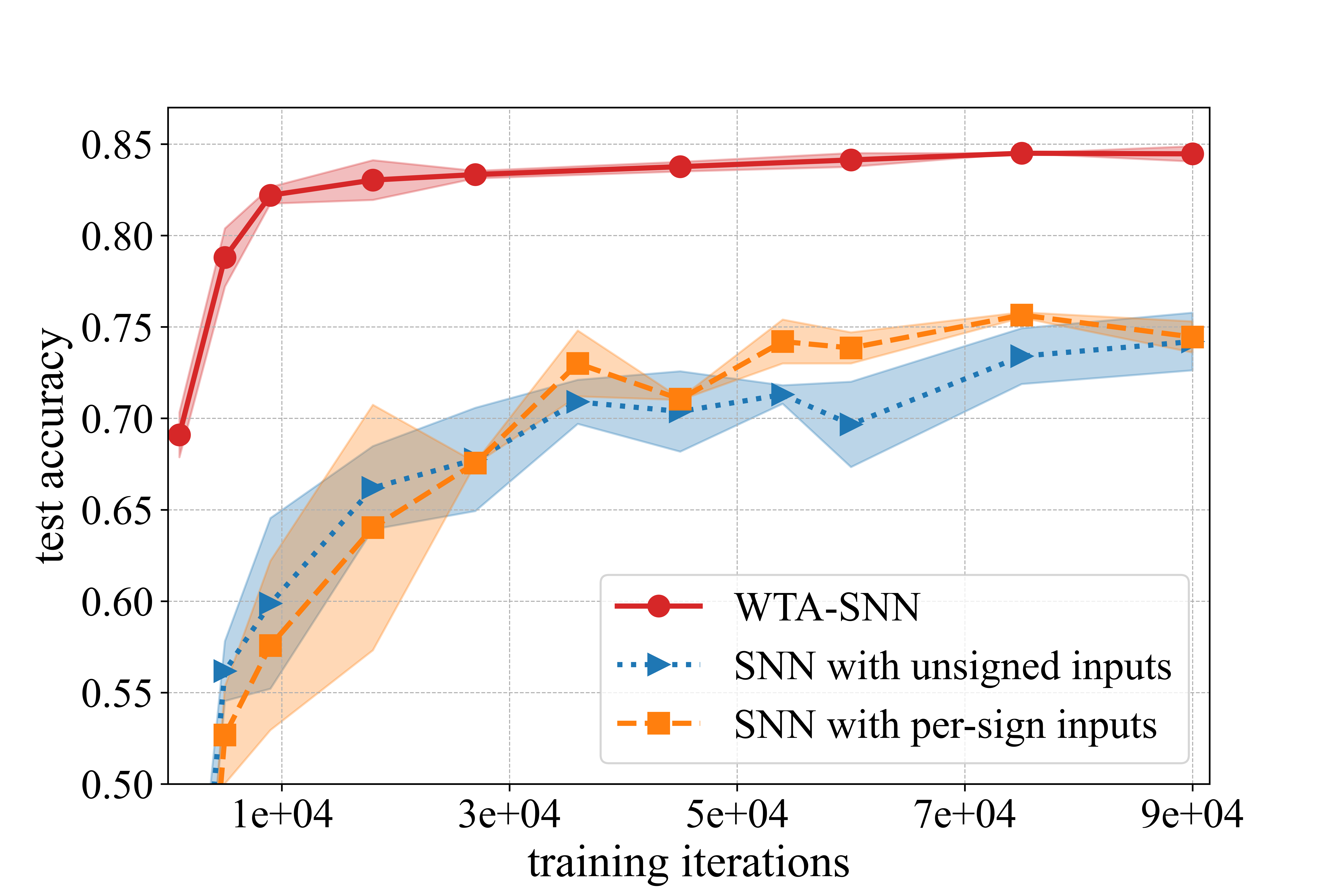}}
\vspace{-0.3cm}
\caption{Test accuracy of binary SNNs with unsigned or per-sign inputs, and WTA-SNN, all trained via {\VOWEL} for the MNIST-DVS dataset. Shaded areas represent confidence intervals equal to one standard deviation.}
\label{fig:multi_valued}
\end{center}
\vspace{-0.6cm}
\end{figure}

We now turn to the comparison of WTA-SNNs trained with {\VOWEL} with the mentioned baseline binary SNNs. Fig.~\ref{fig:multi_valued} shows the classification accuracy in the test set as a function of the training samples. The WTA-SNN trained with {\VOWEL} is seen to outperform both conventional binary SNN solutions, which were also trained with {\VOWEL} (with $C = 1$). In line with previous results \cite{zhao2014feedforward, henderson2015spike}, the use of sign information is only marginally useful for a conventional SNN. This demonstrates the unique capability of WTA circuits to process information encoded both in the spikes' timings and in their associated values, here the signs.



\subsection{DVSGesture Classification}
Having elaborated on the comparison with conventional SNNs trained with {\VOWEL}, we now compare the performance of the proposed WTA-SNN system with a binary SNN trained using a state-of-the-art method known as {\DECOLLE} \cite{kaiser2018decolle}. {\DECOLLE} assumes a layered architecture, and it carries out credit assignment by defining a random layer-wise rate-encoded target that depends on the current average spiking rate in the layer. 
For {\DECOLLE}, the $2H$ hidden neurons are equally divided into two layers. We consider a challenging gesture classification task based on the DVSGesture dataset \cite{amir2017dvsgesture}. For {\VOWEL}, we consider a network with up to $H = 256$ hidden WTA circuits. An equal or larger number of spiking units is considered for {\DECOLLE}. In order to explore the trade-off between test classification accuracy and complexity, we have evaluated the performance of {\DECOLLE} under different sampling rates from $1$ ms to $20$ ms, with the latter corresponding to the sampling rate used for {\VOWEL}.

A summary of the performance comparison between {\DECOLLE} and {\VOWEL} is provided in Table~\ref{tab:comparison-gesture}. 
It is first noted that, owing to the more significant complexity of the task under study with respect to the MNIST-DVS task, the accuracy levels are generally lower. We also observe that the performance of {\DECOLLE} degrades quickly with decreasing number of hidden neurons and sampling rates. 
In particular, with a sampling period of $20$ ms and $H=256$ (i.e., two layers of 256 neurons), the accuracy drops to $26.38 \%$.  This is likely due to the layered architecture with local rate-encoded credit assignment used by {\DECOLLE} and to the importance for the DVSGesture dataset to extract information encoded in spatio-temporal patterns. 
In contrast, WTA-SNNs trained with {\VOWEL} provide more robust accuracy levels even with only $H = 64$ circuits. We attribute this capacity to operate with a smaller number of neurons and coarser sampling rates to the power of WTA-SNNs to directly distinguish spatio-temporal patterns encoded in the values of the spikes. 

\section{Related Works} \label{sec:related}

Probabilistic models for SNNs have been mostly proposed as solutions to implement approximate Bayesian inference through sampling and rate encoding in Boltzmann Machines (BM) \cite{buesing2011neural, neftci2014event}. 
We recall that BMs are undirected generative models modeling correlated (static) scalar variables. The approach, known as {\em neural sampling}, was pioneered in \cite{buesing2011neural} for a two-layer network consisting of a visible layer of spiking neurons and a latent WTA circuit. 
We note that the BM can be characterized as a special case of \eqref{eq:binary-ind-prob} with $K=1$ and with unit synaptic memory, where the unitary memory corresponds to the values sampled at previous instant for Gibbs sampling and information is encoded in the average spiking rate of the neurons. Learning is carried out via a Monte Carlo-based EM yielding local update rules or via contrastive divergence \cite{neftci2014event}, while inference of the latent variables is carried out via Gibbs sampling on the BM.


While BMs are undirected graphical models, 
inference on more general directed models defined on a tree is studied in \cite{guo2017hierarchical}. This work uses randomness in the spiking behavior of WTA circuits arranged on a tree to implement Bayesian inference in trees of categorical variables through mean-field variational inference. Learning is carried out via variational EM, resulting in local update rules based on STDP. Generative models based on probabilistic SNNs are studied in \cite{mostafa2018learning}, which considers directed networks of WTA circuits with noisy synapses. The network is trained via variational EM based on the reparameterization trick and diffusion approximation (see also \cite{neftci2014event}).

\begin{table}[t]
\caption{Test accuracy of {\VOWEL} and {\DECOLLE} on DVSGesture}
\label{tab:comparison-gesture}
\begin{center}
\begin{small}
\begin{sc}
\begin{tabular}{lcccr}
\toprule
Model & Period & $H$ & Accuracy \\
\midrule
 & 1 ms & $1024$ &  $61.42 \pm 2.92$\%  \\
 & 1 ms & $512$ &  $57.75 \pm 3.22$\%  \\
{\DECOLLE} \cite{kaiser2018decolle} & 1 ms & $512$ &  $56.42 \pm 2.03$\%  \\
 & 10 ms & $256$ & $34.72 \pm 0.75$\% \\
 & 20 ms & $256$ & $26.38 \pm 0.28$\% \\
\midrule
& 20 ms & 256 & $60.26 \pm 0.91\%$ \\
{\VOWEL} & 20 ms & 128 & $57.96 \pm 0.11\%$ \\
 & 20 ms & 64 & $57.89 \pm 0.35\%$ \\
\bottomrule
\end{tabular}
\end{sc}
\end{small}
\end{center}
\vspace{-0.5cm}
\end{table}

All the works reviewed above use probabilistic SNNs to approximate inference or data generation in networks of correlated categorical variables by means of sampling and rate encoding. 
These solutions are hence unable to effectively process time encoded data in which information is carried by 
spatio-temporal patterns. 
In contrast, learning rules for probabilistic SNNs were introduced in \cite{brea2013matching} for sequence memorization and in \cite{jimenez2014stochastic, gardner2016supervised} for sequence-to-sequence mapping. 
Finally, multi-valued spikes (deterministic) models are studied in \cite{xu2019boosting} to reduce energy consumption of pre-designed SNNs via time compression.

\section{Conclusions and Discussion} \label{sec:discussion}

In this paper, we have proposed the first general local and online learning rule for WTA-SNNs -- referred to as {\VOWEL} -- that applies to any topology. {\VOWEL} only requires a scalar feedback signal for the update of the weights of hidden circuits. It is derived starting from first principles as the maximization of a likelihood metric, and it encompasses KL-based regularization and baseline control variates. Experiments have demonstrated that WTA-SNN systems trained with {\VOWEL} can outperform state-of-the-art methods based on conventional SNNs for the solution of tasks in which information is encoded both in spatio-temporal patterns and in numerical values attached to spikes. The advantages are particularly evident for resource-constrained settings with a small number of hidden neural units.

While this work considered the maximization of a discounted log-likelihood learning criterion, similar rules can be derived by replacing the log-likelihood with other reward functions. For example, in \cite{bleema2018fts}, a probabilistic WTA-SNN can be used as a stochastic policy in reinforcement learning. 
Other generalizations include Bayesian learning via Langevin dynamics \cite{kappel15:synaptic, jang19:def}; and the optimization of tighter lower bounds on the likelihood \cite{burda2015importance, mnih2016variational}.

\bibliographystyle{IEEEtran}
\bibliography{ref}

\newpage
\onecolumn
\appendices
\section{Binary SNNs}

\subsection{Methods: Training GLM-based Binary SNNs} 

{\bf Online Local Variational EM-based Rule.} 
An online local update rule for the model parameters can be obtained by considering the maximization of this criterion $L_{\bmx_{\leq t}}(\bmtheta)$ via stochastic gradient descent \cite{jimenez2014stochastic, brea2013matching, jang19:spm}. 
To this end, at each time $t$, each hidden neuron $i \in \set{H}$ generates a sample $h_{i,t} \sim p_{\bmtheta_i}(h_{i,t}|u_{i,t})$ from \eqref{eq:binary-ind-prob}. Then, for visible neurons $i \in \set{X}$, the synaptic weights are updated as
\begin{subequations} \label{eq:binary-update}
\begin{eqnarray} \label{eq:binary-update-vis}
    \Delta w_{j,i}^{(k)} = \eta \cdot \Big\langle \underbrace{ \big( x_{i,t} - \sigma(u_{i,t}) \big)}_{\text{post}_i} \cdot \underbrace{ \overra{s}_{j,t-1}^{(k)}}_{\text{pre}_j^{(k)}} \Big\rangle_{\gamma},
\end{eqnarray}
while, for hidden neurons $i \in \set{H}$, we have the update
\begin{eqnarray} \label{eq:binary-update-hid}
    \Delta w_{j,i}^{(k)} = \eta \cdot \Big\langle \underbrace{\ell_{\bmtheta^\text{X},t}}_{\text{reward}} \cdot~ \big\langle \underbrace{ \big( h_{i,t} - \sigma(u_{i,t}) \big)}_{\text{post}_i} \cdot  \underbrace{ \overra{s}_{j,t-1}^{(k)}}_{\text{pre}_j^{(k)}} \big\rangle_{\kappa} \Big\rangle_{\gamma}
\end{eqnarray} 
\end{subequations}
for some time-average coefficient $0<\kappa<1$, and we have defined the reward, or learning signal, at time $t$ as $\sum_{i \in \set{X}} \bar{H}\big( x_{i,t}, \sigma(u_{i,t}) \big)$. Similar expressions apply for the other model parameters (see Algorithm~$2$ in \cite{jang19:spm}). We have generalized this rule in Sec.~\ref{sec:vem-wta}, which provides the expressions for all updates. We also discuss there the relationship of rule \eqref{eq:binary-update} with related algorithms derived from deterministic models.

\section{WTA-SNNs and Online Local VEM-based Learning Rule for WTA-SNNs.}

\subsection{Details on Probabilistic Models} 

\begin{figure*}[ht]
\centering
\subfigure[]{
\includegraphics[height=0.15\columnwidth]{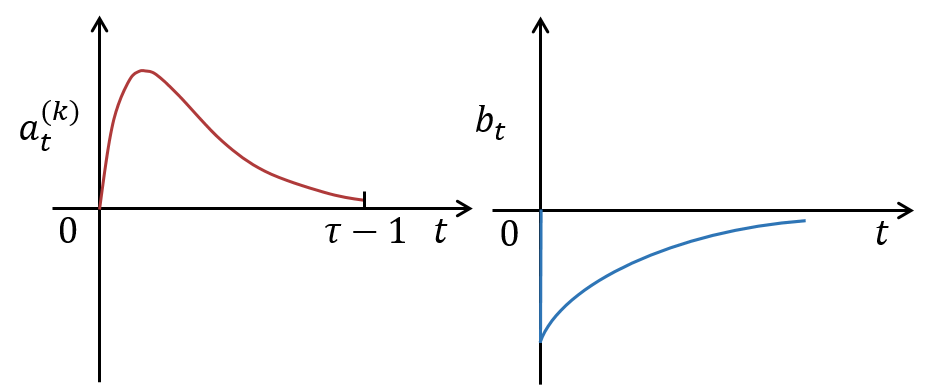}
\label{fig:wta-filter-exp}
}
\subfigure[]{
\includegraphics[height=0.15\columnwidth]{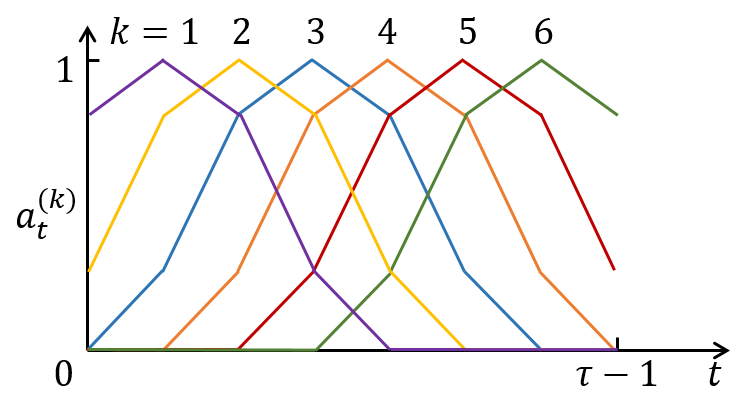}
\label{fig:wta-filter-cosine}
}
\caption{Examples of synaptic, self-memory filters of duration $\tau$: (a) an exponentially decaying synaptic kernel $a_t^{(k)}$ and self-memory kernel $b_t$, and (b) raised cosine raised functions $a_t^{(k)}$ in \cite{pillow2008spatio}.}
\label{fig:wta-filter}
\vspace{-0.15cm}
\end{figure*}

{\bf Spatio-temporal filters.} 
The membrane potential $\bmu_{i,t}$ in \eqref{eq:wta-potential} assumes a set of fixed $K$ finite-duration spatio-temporal filters $\{a_t^{(k)}\}_{k=1}^K$ for synapses and a finite-duration filter $b_t$ for self-memory. For a synapse $(j,i)$ from pre-synaptic circuit $j \in \set{P}_i$ to post-synaptic circuit $i$, the synaptic filters $\{a_t^{(k)}\}_{k=1}^K$, along with the corresponding synaptic weight matrices $\{\bfW_{j,i}^{(k)}\}_{k=1}^K$, determine the temporal sensitivity of the synapse. Specifically, the contribution of the $k$th synaptic trace $\overra{\bms}_{j,t-1}^{(k)}$ is multiplied by a learnable weight matrix $\bfW_{j,i}^{(k)}$, where the $(c,c')$th element indicates the contribution of the past incoming signals from circuit $j$ with value $c'$ to the propensity of circuit $i$ to emit a spike of value $c$. In a similar manner, the feedback, or self-memory, filter $b_t$, along with the feedback weight matrix $\bfW_i$ which describes the feedback contribution of the past spiking signals of the circuit itself, determine the temporal response of a circuit $i$ to its own spikes. When the filter $a_t^{(k)}$ is of finite duration $\tau$, computing the filtered trace $\overra{\bms}_{i,t}^{(k)}$ requires keeping track of the window $\{\bms_{i,t}, \bms_{i,t-1}, \ldots, \bms_{i,t-(\tau-1)} \}$ of prior signals. An example is given by exponentially decaying function of the form $a_t^{(k)} = \exp(-t/\tau_1) - \exp(-t/\tau_2)$ for $t \leq \tau-1$ and zero otherwise, with time constants $\tau_1, \tau_2$. An example of negative feedback filter $b_t = -\exp(-t/\tau_3)$ with time constant $\tau_3$ models the refractory period upon the emission of a spike, with the time constant $\tau_3$ determining the refractory period duration, see Fig.~\ref{fig:wta-filter-exp}. Example of filter functions include raised cosine functions with different synaptic delays, see Fig.~\ref{fig:wta-filter-cosine}.

{\bf Probabilistic GLM model of WTA-SNNs.} 
The joint probability of the signals $\bms_{\leq T} = (\bms_1, \ldots, \bms_T)$ emitted by all circuits in the WTA-SNN up to time $T$ can be written using the chain rule as
\begin{align} \label{eq:wta-joint}
    p_{\bmtheta}(\bms_{\leq T}) = \prod_{t=1}^T p_{\bmtheta}(\bms_t | \bms_{\leq t-1}) = \prod_{t=1}^T \prod_{i \in \set{V}} p_{\bmtheta_i}(\bms_{i,t}|\bms_{\leq t-1}),
\end{align}
where $\bmtheta = \{\bmtheta_i\}_{i \in \set{V}}$ is a set of model parameters. Under the considered GLM model \eqref{eq:wta-onehot}-\eqref{eq:wta-potential}, the probability of circuit $i$ to output a spike at unit $c$ for time $t$ is 
\begin{align*}
    p_{\bmtheta_i}(\bms_{i,t} = \bfe_c | \bmu_{i,t}) = \sigma_c(\bmu_{i,t}) = \frac{ \exp\big( \bmu_{i,t}^\top \bfe_c \big)}{ 1 + \sum_{c'=1}^C \exp\big( \bmu_{i,t}^\top \bfe_{c'} \big)}
\end{align*}
for $c=1,\ldots, C$, while the probability of not spiking is given as
\begin{align*}
    p_{\bmtheta_i}(\bms_{i,t} = {\bm 0} | \bmu_{i,t}) = \frac{1}{ 1 + \sum_{c'=1}^C \exp\big( \bmu_{i,t}^\top \bfe_{c'} \big)}.
\end{align*}
Defining the vector function $\bmsigma(\bmu_{i,t}) = [\sigma_1(\bmu_{i,t}), \ldots, \sigma_C(\bmu_{i,t})]^\top$, it can be rewritten in a more compact form as
\begin{align}
    p_{\bmtheta_i}(\bms_{i,t}|\bmu_{i,t}) =
    \begin{cases}
    1 - \sum_{c=1}^C \sigma_c(\bmu_{i,t}), ~~~~~~\text{if}~~ \bms_{i,t} = {\bm 0}, \\
\bms_{i,t}^\top \bmsigma(\bmu_{i,t}), \qquad \quad ~~~~~~~~\text{if}~~ \bms_{i,t} = \bfe_c ~~\text{for}~~ c \in \{1,2,\ldots,C\}.
    \end{cases}
\end{align}
Moreover, it follows that the log-probability corresponds to the categorical negative cross-entropy 
\begin{align} \label{eq:wta-ind-ce2}
    \log p_{\bmtheta_i}(\bms_{i,t}|\bmu_{i,t}) = \sum_{c=1}^C {\bm 1}_{\{\bms_{i,t}=\bfe_c\}} \cdot \log \sigma_c(\bmu_{i,t}) + {\bm 1}_{\{\bms_{i,t} = {\bm 0}\}} \cdot \log \Big( 1-\sum_{c'=1}^C \sigma_{c'}(\bmu_{i,t}) \Big) = \bar{H}\big(\bms_{i,t}, \bmsigma(\bmu_{i,t})\big).
\end{align}
We note that the categorical negative cross-entropy of circuit $i$ depends on the local model parameter $\bmtheta_i$ through the membrane potential $\bmu_{i,t}$, which we omit for notational simplicity. 

{\bf Derivation of the gradient $\grad_{\bmtheta_i} \log p_{\bmtheta_i}(\bms_{i,t}|\bmu_{i,t})$ in \eqref{eq:wta-ll-grad}.} 
From the chain rule, we have that 
\begin{align*}
    \grad_{\bmtheta_i} \log p_{\bmtheta_i}(\bms_{i,t}|\bmu_{i,t}) &= \grad_{\bmu_{i,t}} \bar{H}\big( \bms_{i,t}, \bmsigma(\bmu_{i,t})\big) \cdot \grad_{\bmtheta_i} \bmu_{i,t} = \Big( \bms_{i,t} - \bmsigma(\bmu_{i,t}) \Big) \cdot \grad_{\bmtheta_i} \bmu_{i,t},
\end{align*}
where the partial derivative $\grad_{\bmu_{i,t}} \bar{H}\big( \bms_{i,t}, \bmsigma(\bmu_{i,t})\big)$ is computed in a similar manner to the gradient of the standard categorical cross-entropy. Specifically, for each $c$th entry of the membrane potential, say $u_{i,t}^{(c)}$, it follows that  
\begin{align*}
    \grad_{u_{i,t}^{(c)}} \bar{H}\big( \bms_{i,t}, \bmsigma(\bmu_{i,t}) \big) = {\bm 1}_{\{\bms_{i,t} = \bfe_{c}\}} - \sigma_c(\bmu_{i,t})
\end{align*}
by plugging the following fact into the gradient of \eqref{eq:wta-ind-ce2}
\begin{align*}
    \frac{\partial \sigma_c(\bmu_{i,t})}{\partial u_{i,t}^{(c)}} = 
    \sigma_c(\bmu_{i,t}) \cdot \Big(1-\sigma_c(\bmu_{i,t})\Big), \quad \text{and} \quad  
    \frac{\partial \sigma_{c'}(\bmu_{i,t})}{\partial u_{i,t}^{(c)}} = 
    - \sigma_c(\bmu_{i,t}) \cdot \sigma_{c'}(\bmu_{i,t}) \quad \text{for}~~ c' \neq c.
\end{align*}
The partial derivative $\grad_{\bmtheta_i} \bmu_{i,t}$ is directly computed from the \eqref{eq:wta-potential} as
\begin{align*}
    \grad_{\bmvartheta_i} \bmu_{i,t} = 1, \quad
    \grad_{\bfW_{j,i}^{(k)}} \bmu_{i,t} = \overra{\bms}_{j,t-1}^{(k)}, \quad
    \grad_{\bfW_i} \bmu_{i,t} = \overla{\bms}_{i,t-1}.
\end{align*}
As a result, we have 
\begin{align*}
    \grad_{\bmvartheta_i} \log p_{\bmtheta_i}(\bms_{i,t}|\bmu_{i,t}) &= \bms_{i,t} - \bmsigma(\bmu_{i,t}), \cr
    \grad_{\bfW_{j,i}^{(k)}} \log p_{\bmtheta_i}(\bms_{i,t}|\bmu_{i,t}) &= \big( \bms_{i,t} - \bmsigma(\bmu_{i,t}) \big) \big(\overra{\bms}_{j,t-1}^{(k)}\big)^\top, ~~\text{for}~~ k = 1,\ldots,K, \cr 
    \grad_{\bfW_i} \log p_{\bmtheta_i}(\bms_{i,t}|\bmu_{i,t}) &= \big(\bms_{i,t} - \bmu_{i,t}\big) \overla{\bms}_{i,t-1}^\top.
\end{align*}

\subsection{Variational EM for WTA-SNNs} 

{\bf Evidence of Lower Bound (ELBO) for WTA-SNNs.} 
Learning can be carried out by maximizing the likelihood (ML) that a subset $\set{X}$ of visible circuits outputs desired spiking signals $\bmx_{\leq T}$ in response to given inputs. Mathematically, we can write the ML problem as $\max_{\bmtheta} \log p_{\bmtheta}(\bmx_{\leq T})$, where the stochastic spiking signals $\bmh_{\leq T}$ of hidden circuits in the complementary set $\set{H}$ have to be averaged over in order to evaluate the marginal log-likelihood $\log p_{\bmtheta}(\bmx_{\leq T}) = \log \sum_{\bmh_{\leq T}} p_{\bmtheta}(\bmx_{\leq T}, \bmh_{\leq T})$. To tackle the problem, variational EM maximizes the evidence of lower bound (ELBO)  
\begin{align*}
    L_{\bmx_{\leq T}}(\bmtheta) = \mathbb{E}_{q(\bmh_{\leq T})} \Big[ \log \frac{p_{\bmtheta}(\bmx_{\leq T}, \bmh_{\leq T})}{q(\bmh_{\leq T})} \Big]
\end{align*}
over both the model parameters $\bmtheta$ and the variational posterior distribution $q(\bmh_{\leq T}) = \prod_{t=1}^T q(\bmh_t | \bmh_{\leq t-1})$.

As in \eqref{eq:binary-caually-cond} for conventional SNNs, we choose the variational distribution to equal the causally conditioned distribution of the hidden circuits given the visible circuits 
\begin{align*}
    q(\bmh_{\leq T}) = p_{\bmtheta^\text{H}}(\bmh_{\leq T} || \bmx_{\leq T-1}) = \prod_{t=1}^T \prod_{i \in \set{H}} p_{\bmtheta_i}(\bmh_{i,t} | \bmu_{i,t}).
\end{align*}
From \eqref{eq:wta-joint}, this choice results the ratio in the ELBO $L_{\bmx_{\leq T}}(\bmtheta)$ equal to the categorical negative cross-entropy loss for the visible circuits, i.e., 
\begin{align*}
    \log \frac{p_{\bmtheta}(\bmx_{\leq T}, \bmh_{\leq T})}{q(\bmh_{\leq T})} = \log p_{\bmtheta^\text{X}}(\bmx_{\leq T} || \bmh_{\leq T-1}) = \sum_{t=1}^T \sum_{i \in \set{X}} \bar{H}\big( \bmx_{i,t}, \bmsigma(\bmu_{i,t}) \big).
\end{align*}
We also note that the ratio corresponds to the causally conditioned distribution of the visible circuits given the hidden circuits. The resulting ELBO for WTA-SNNs can be written as 
\begin{align*}
    L_{\bmx_{\leq T}}(\bmtheta) = \mathbb{E}_{p_{\bmtheta^\text{H}}(\bmh_{\leq T} || \bmx_{\leq T-1})} \Big[ \sum_{t=1}^T \sum_{i \in \set{X}} \bar{H}\big( \bmx_{i,t}, \bmsigma(\bmu_{i,t})\big) \Big].
\end{align*}

{\bf ELBO with Variational Regularization for WTA-SNNs.} 
Following the approach of bounded rationality \cite{leibfried2015reward} regularization, sparsity-inducing variational regularization can be applied by adding a KL divergence term between the hidden circuits' distribution $p_{\bmtheta^\text{H}}(\bmh_{\leq T} || \bmx_{\leq T-1})$ and a baseline distribution $r(\bmh_{\leq T}) = \prod_{t=1}^T \prod_{i \in \set{H}} r(\bmh_{i,t})$ with the desired level of sparsity $r(\bmh_{i,t})$ of each hidden circuit $i$ at time $t$. Specifically, with a regularization coefficient $\alpha > 0$, the learning criterion is given as 
\begin{align*}
    &L_{\bmx_{\leq T}}(\bmtheta) - \alpha \cdot \text{KL}\big( p_{\bmtheta^\text{H}}(\bmh_{\leq T} || \bmx_{\leq T-1}) || r(\bmh_{\leq T}) \big) \cr 
    &\quad = \mathbb{E}_{p_{\bmtheta^\text{H}}(\bmh_{\leq T} || \bmx_{\leq T-1})} \Big[ \log p_{\bmtheta^\text{X}}(\bmx_{\leq T} || \bmh_{\leq T-1}) \Big] - \alpha \cdot \mathbb{E}_{p_{\bmtheta^\text{H}}(\bmh_{\leq T} || \bmx_{\leq T-1})} \Big[ \log \frac{p_{\bmtheta^\text{H}}(\bmh_{\leq T} || \bmx_{\leq T-1})}{r(\bmh_{\leq T})} \Big] \cr 
    &\quad = \mathbb{E}_{p_{\bmtheta^\text{H}}(\bmh_{\leq T} || \bmx_{\leq T-1})} \bigg[ \sum_{t=1}^T \bigg( \underbrace{ \sum_{i \in \set{X}} \bar{H}\big( \bmx_{i,t},\bmsigma(\bmu_{i,t})\big) - \alpha \cdot \sum_{i \in \set{H}} \Big( \bar{H}\big( \bmh_{i,t},\bmsigma(\bmu_{i,t})\big) - \log r(\bmh_{i,t}) \Big) }_{:=~ \ell_{\bmtheta, t}:~ \text{reward with regularization at time $t$}} \bigg) \bigg] := L_{\bmx_{\leq T}}^\text{reg}(\bmtheta).
\end{align*}
To obtain online learning rules, we aim at maximizing a discounted version of the lower bound $L_{\bmx_{\leq T}}^\text{reg}(\bmtheta)$ at each time $t$ as
\begin{align*}
    L_{\bmx_{\leq t}}^\text{reg}(\bmtheta) = \mathbb{E}_{p_{\bmtheta^\text{H}}(\bmh_{\leq t} || \bmx_{\leq t-1})} \bigg[ \sum_{t'=0}^{t-1} \gamma^{t'} \ell_{\bmtheta, t-t'} \bigg],
\end{align*}
where $0 < \gamma < 1$ is a discount factor.

{\bf Details on temporal average $\langle \cdot \rangle$.} 
Before to derive an online learning rule maximizing the lower bound $L_{\bmx_{\leq t}}^\text{reg}(\bmtheta)$, we first note that the temporal average $\langle f_t \rangle_\gamma$ of a time sequence $\{ f_t \}_{t \geq 1}$ with a constant $\gamma \in (0,1)$ equals to the discounted sum of the time sequence $\sum_{t'=0}^{t-1} \gamma^{t'} f_{t-t'}$. With $\langle f_0 \rangle_\gamma = 0$, from recursive computations of the temporal average operator, we have 
\begin{align*}
    \langle f_t \rangle_\gamma &= \gamma \cdot \langle f_{t-1} \rangle_\gamma + f_t 
    = \gamma \cdot \big( \gamma \cdot \langle f_{t-2} \rangle_\gamma + f_{t-1} \big) + f_t  
    = \ldots \cr
    &= f_t + \gamma f_{t-1} + \gamma^2 f_{t-2} + \cdots + \gamma^{t-1} f_1 
    = \sum_{t'=0}^{t-1} \gamma^{t'} f_{t-t'}.
\end{align*}
Using the temporal average, the lower bound is obtained as $L_{\bmx_{\leq t}}^\text{reg}(\bmtheta) = \mathbb{E}_{p_{\bmtheta^\text{H}}(\bmh_{\leq t} || \bmx_{\leq t-1})} \Big[ \big\langle \ell_{\bmtheta,t} \big\rangle_\gamma \Big]$. Finally, we note that the summation of a time sequence $\{f_t\}_{t \geq 1}$ over time, i.e., $\sum_{t'=1}^t f_{t'}$, can be approximated with the temporal average $\langle f_t \rangle_\kappa$ with some constant $\kappa \in (0,1)$, where the constant $\kappa$ close to $1$ allows to get a better proxy.

{\bf Derivation of the gradient $\grad_{\bmtheta} L_{\bmx_{\leq t}}^\text{reg}(\bmtheta)$. }
We compute the gradient of the objective $L_{\bmx_{\leq t}}^\text{reg}(\bmtheta)$ with respect to $\bmtheta$ as
\begin{align*}
    \grad_{\bmtheta} L_{\bmx_{\leq t}}^\text{reg}(\bmtheta) &\stackrel{(a)}{=} \mathbb{E}_{p_{\bmtheta^\text{H}}(\bmh_{\leq t} || \bmx_{\leq t-1})} \Big[ \Big\langle \grad_{\bmtheta} \ell_{\bmtheta,t} \Big\rangle_\gamma + \Big\langle \ell_{\bmtheta,t} \Big\rangle_\gamma \cdot \grad_{\bmtheta} \log p_{\bmtheta^\text{H}}(\bmh_{\leq t} || \bmx_{\leq t-1}) \Big] \cr 
    &= \mathbb{E}_{p_{\bmtheta^\text{H}}(\bmh_{\leq t} || \bmx_{\leq t-1})} \bigg[ \Big\langle \grad_{\bmtheta} \ell_{\bmtheta,t} \Big\rangle_\gamma + \sum_{t'=1}^t \gamma^{t-t'} \ell_{\bmtheta,t'} \cdot \sum_{t'=1}^t \sum_{i \in \set{H}} \grad_{\bmtheta} \bar{H}\big( \bmh_{i,t'}, \bmsigma(\bmu_{i,t'})\big) \bigg] \cr 
    &\stackrel{(b)}{=} \mathbb{E}_{p_{\bmtheta^\text{H}}(\bmh_{\leq t} || \bmx_{\leq t-1})} \bigg[ \Big\langle \grad_{\bmtheta} \ell_{\bmtheta,t} \Big\rangle_\gamma + \sum_{t'=1}^t \Big( \gamma^{t-t'} \ell_{\bmtheta,t'} \cdot \underbrace{ \sum_{t''=1}^{t'} \sum_{i \in \set{H}} \grad_{\bmtheta} \bar{H}\big( \bmh_{i,t''},\bmsigma(\bmu_{i,t''})\big) }_{\approx~ \big\langle \sum_{i \in \set{H}} \grad_{\bmtheta} \bar{H}\big( \bmh_{i,t'}, \bmsigma(\bmu_{i,t'})\big) \big\rangle_\kappa } \Big) \bigg] \cr 
    &\stackrel{(c)}{=} \mathbb{E}_{p_{\bmtheta^\text{H}}(\bmh_{\leq t} || \bmx_{\leq t-1})} \bigg[ \Big\langle \grad_{\bmtheta} \ell_{\bmtheta,t} \Big\rangle_\gamma + \Big\langle \ell_{\bmtheta,t} \cdot \Big\langle \sum_{i \in \set{H}} \grad_{\bmtheta} \bar{H}\big( \bmh_{i,t}, \bmsigma(\bmu_{i,t})\big) \Big\rangle_\kappa \Big\rangle_\gamma \bigg].
\end{align*}
The equality (a) follows directly from an application of the score function, or REINFORCE that uses the likelihood ratio $\grad_{\bmtheta} \log p_{\bmtheta^\text{H}}(\bmh_{\leq t} || \bmx_{\leq t-1}) = \grad_{\bmtheta} p_{\bmtheta^\text{H}}(\bmh_{\leq t} || \bmx_{\leq t-1}) / p_{\bmtheta^\text{H}}(\bmh_{\leq t} || \bmx_{\leq t-1})$. Recalling that under mild conditions on a parametric distribution $p_\theta(x)$ we have 
\begin{align*}
    \mathbb{E}_{p_{\theta}(x)}[ \grad_{\theta} \log p_{\theta}(x)] = \sum_x p_{\theta}(x) \grad_{\theta} \log p_{\theta}(x) = \sum_x \grad_{\theta} p_{\theta}(x) = \grad_{\theta} \sum_x p_{\theta}(x) = \grad_{\theta} 1 = 0,
\end{align*}
we can simplify the gradient through the equality (b) as in \cite{peters2008reinforcement}. In (c), the summation over time up to $t$ is estimated using temporal average operator with a constant $\kappa \in (0,1)$. In detail, for visible circuit $i \in \set{X}$, the gradient equals to 
\begin{subequations} \label{eq:wta-grad}
\begin{align} \label{eq:wta-grad-vis}
    \grad_{\bmtheta_i} L_{\bmx_{\leq t}}^\text{reg}(\bmtheta) &= \mathbb{E}_{p_{\bmtheta^\text{H}}(\bmh_{\leq t} || \bmx_{\leq t-1})} \Big[ \Big\langle \grad_{\bmtheta_i} \ell_{\bmtheta,t} \Big\rangle_\gamma \Big] 
    = \mathbb{E}_{p_{\bmtheta^\text{H}}(\bmh_{\leq t} || \bmx_{\leq t-1})} \Big[ \Big\langle \grad_{\bmtheta_i} \bar{H}\big( \bmx_{i,t}, \bmsigma(\bmu_{i,t})\big) \Big\rangle_\gamma \Big],
\end{align}
while for hidden circuit $i \in \set{H}$, we have
\begin{align} \label{eq:wta-grad-hid}
    \grad_{\bmtheta_i} L_{\bmx_{\leq t}}^\text{reg}(\bmtheta) &= \mathbb{E}_{p_{\bmtheta^\text{H}}(\bmh_{\leq t} || \bmx_{\leq t-1})} \Big[ \Big\langle -\alpha \cdot \grad_{\bmtheta_i} \bar{H}\big( \bmh_{i,t},\bmsigma(\bmu_{i,t})\big) \Big\rangle_\gamma + \Big\langle \ell_{\bmtheta,t} \cdot \Big\langle  \grad_{\bmtheta_i} \bar{H}\big( \bmh_{i,t}, \bmsigma(\bmu_{i,t})\big) \Big\rangle_\kappa \Big\rangle_\gamma \Big] \cr 
    &= \mathbb{E}_{p_{\bmtheta^\text{H}}(\bmh_{\leq t} || \bmx_{\leq t-1})} \Big[ \Big\langle \ell_{\bmtheta,t} \cdot \Big\langle \grad_{\bmtheta_i} \bar{H}\big( \bmh_{i,t}, \bmsigma(\bmu_{i,t})\big) \Big\rangle_\kappa \Big\rangle_\gamma \Big].
\end{align}
\end{subequations}

{\bf Monte Carlo estimate of the gradient.} 
An Monte Carlo estimate of the gradient \eqref{eq:wta-grad} can be obtained by drawing a single sample $\bmh_{\leq t}$ of hidden circuits from the causally conditioned distribution $p_{\bmtheta^\text{H}}(\bmh_{\leq t} || \bmx_{\leq t-1})$ and evaluating 
\begin{subequations} \label{eq:wta-grad-mc}
\begin{align}
    \grad_{\bmtheta_i} \hat{L}_{\bmx_{\leq t}}^\text{reg}(\bmtheta) &= \Big\langle \grad_{\bmtheta_i} \bar{H}\big( \bmx_{i,t}, \bmsigma(\bmu_{i,t})\big) \Big\rangle_\gamma, ~~ i \in \set{X}, \label{eq:wta-grad-mc-vis} \\ 
    \grad_{\bmtheta_i} \hat{L}_{\bmx_{\leq t}}^\text{reg}(\bmtheta) &= \Big\langle \big( \ell_{\bmtheta,t} - \bmb_{i,t} \big) \cdot \underbrace{ \big\langle \grad_{\bmtheta_i} \bar{H}\big( \bmh_{i,t},\bmsigma(\bmu_{i,t})\big) \big\rangle_\kappa }_{:=~ \bme_{i,t}} \Big\rangle_\gamma, ~~ i \in \set{H}. \label{eq:wta-grad-mc-hid}
\end{align}
\end{subequations}
We have introduced baseline, or control variates, signals $\bmb_{i,t}$ as means to reduce the variance of the gradient estimator. Following the approach in \cite{peters2008reinforcement}, an optimized baseline at each time $t$ that minimizes a bound on the variance of the gradient can be selected as 
\begin{equation*} 
    \bmb_{i,t} = \frac{ \mathbb{E}\Big[ \ell_{\bmtheta,t} \cdot \bme_{i,t}^2 \Big]}{ \mathbb{E} \Big[ \bme_{i,t}^2 \Big]}.
\end{equation*}
In the proposed implementation in {\VOWEL}, the expectations are approximated using averages over time, i.e., 
\begin{equation*} 
    \bmb_{i,t} = \frac{\big\langle \ell_{\bmtheta,t} \cdot \bme_{i,t}^2 \big\rangle_{\kappa_b}}{\big\langle \bme_{i,t}^2 \big\rangle_{\kappa_b}},
\end{equation*}
with some constant $\kappa_b \in (0,1)$.

\subsection{Detailed Algorithm of {\VOWEL}}

To elaborate the proposed learning rule {\VOWEL}, at each time $t$, spiking signal $\bmh_{i,t}$ of each hidden circuit $i \in \set{H}$ is locally generated from the current model $\bmh_{i,t} \sim p_{\bmtheta_i}(\bmh_{i,t} | \bmu_{i,t})$ and the global reward $\ell_{\bmtheta,t}$ in \eqref{eq:wta-ls} is computed by a central processor by collecting the membrane potentials $\bmu_{i,t}$ from all circuits $i \in \set{V}$. Finally, as detailed in Algorithm~\ref{alg:vowel}, for each circuit $i$, {\VOWEL} updates the local model parameters $\bmtheta_i$ via stochastic gradient descent as $\bmtheta_i \leftarrow \bmtheta_i + \eta \cdot \grad_{\bmtheta_i} \hat{L}_{\bmx_{\leq t}}^\text{reg}(\bmtheta)$ in the direction of a stochastic Monte Carlo estimate \eqref{eq:wta-grad-mc}, with learning rate $\eta$ as in \eqref{eq:wta-online-update}.

\begin{algorithm}[th]
   \caption{{\VOWEL}: Online Learning Rule of WTA-SNNs via Variational EM}
   \label{alg:vowel}
\begin{algorithmic}
   \STATE {\bfseries Input:} data $\bmx_{\leq t}$, discount factor $\gamma$. time-averaging constants $\kappa, \kappa_b$, learning rate $\eta$, regularization coefficient $\alpha$, and sparsity level $r(\cdot)$
   
   \STATE {\bfseries Output:} learned model parameters $\bmtheta$ 
   \vspace{0.1cm}
   \hrule
   \vspace{0.1cm}
   \STATE {\bf initialize} parameters $\bmtheta$
   
   \FOR{each time $t=1,2,\ldots$}
   \STATE - each WTA circuit $i \in \set{V}$ computes the synaptic traces $\{\overra{\bms}_{i,t-1}^{(k)}\}_{k=1}^K$ and somatic trace $\overla{\bms}_{i,t-1}$ 
   
   \smallskip
   \STATE - each WTA circuit $i \in \set{V}$ computes the membrane potential $\bmu_{i,t}$ from \eqref{eq:wta-potential} based on synaptic traces $\{\{\overra{\bms}_{j,t-1}^{(k)}\}_{k=1}^K\}_{j \in \set{P}_i}$ from pre-synaptic circuits $\set{P}_i$ and somatic trace of itself $\overla{\bms}_{i,t-1}$
   
   \smallskip
   \STATE - each hidden WTA circuit $i \in \set{H}$ outputs a spike of value $c$ with probability $\sigma_c(\bmu_{i,t})$, i.e., $\bmh_{i,t} \sim \bmsigma(\bmu_{i,t})$
   
   \smallskip
   \STATE - a central processor computes the learning reward $\ell_{\bmtheta,t}$ from \eqref{eq:wta-ls}
   
   \smallskip
    \STATE - each visible WTA circuit $i \in \set{X}$ computes the gradient $\grad_{\bmtheta_i} \log p_{\bmtheta_i}(\bmx_{i,t}|\bmu_{i,t})$ from \eqref{eq:wta-ll-grad}
  
  \smallskip
   \STATE - each hidden WTA circuit $i \in \set{H}$ computes the gradient $\grad_{\bmtheta_i} \log p_{\bmtheta_i}(\bmh_{i,t}|\bmu_{i,t})$ from \eqref{eq:wta-ll-grad} and the baseline 
    \begin{align*}
        \bmb_{i,t} = \frac{ \Big\langle \ell_{\bmtheta,t} \cdot \big\langle \grad_{\bmtheta_i} \log p_{\bmtheta_i}(\bmh_{i,t}|\bmu_{i,t}) \big\rangle_\kappa^2 \Big\rangle_{\kappa_b} }{ \Big\langle \big\langle \grad_{\bmtheta_i} \log p_{\bmtheta_i}(\bmh_{i,t}|\bmu_{i,t}) \big\rangle_\kappa^2 \Big\rangle_{\kappa_b} }
    \end{align*}
   
   \STATE - each WTA circuit $i \in \set{V}$ updates local model parameters as 
   \begin{eqnarray*}
       \bmtheta_i \leftarrow \bmtheta_i + \eta \cdot 
       \begin{cases}
       \Big\langle \grad_{\bmtheta_i} \log p_{\bmtheta_i}(\bmx_{i,t}|\bmu_{i,t}) \Big\rangle_\gamma, \qquad \qquad \qquad ~~~~~~\text{if}~ i \in \set{X}, \\
       \Big\langle \big( \ell_{\bmtheta,t} - \bmb_{i,t} \big) \cdot \big\langle \grad_{\bmtheta_i} \log p_{\bmtheta_i}(\bmh_{i,t}|\bmu_{i,t}) \big\rangle_\kappa \Big\rangle_{\gamma}, ~~~\text{if}~ i \in \set{H}.
       \end{cases}
   \end{eqnarray*}
   \ENDFOR
\end{algorithmic}
\end{algorithm}

\newpage

\section{Experiments}

\subsection{MNIST-DVS Dataset}
The MNIST-DVS dataset was obtained in \cite{serrano2015poker} by displaying slowly moving handwritten digits from the MNIST dataset on an LCD monitor and recording the output of a $128 \times 128$ pixel DVS (Dynamic Vision Sensor) camera \cite{lichtsteiner2006128}. The recordings last about 2 s, which amounts to nearly $2,000,000$ samples per example, with a sampling period of $1$ $\mu$s. The camera uses send-on-delta encoding, whereby positive ($+1$) or negative ($-1$) events are recorded if luminosity respectively increases or decreases, and no event ($0$) otherwise. After pre-processing the dataset, each pixel of any input $26 \times 26$ image consists of a discrete-time sequence of positive ($+$), negative ($-$), or non-spike events. Using one-hot encoding for a WTA-SNN as described in \eqref{eq:wta-onehot}, this yields $676$ two-dimensional input sequences, with one input circuit per pixel. When training binary SNNs using unsigned binary inputs, the input sequence is given by $676$ binary sequences that ignore the input's polarities.

Examples are cropped to a length of 2 s and downsampled. Downsampling is done by gathering events happening during a given time period. Events at each pixel during the considered time period are summed, and the sign of the sum is saved as a one-hot vector.

\begin{figure}[ht]
\begin{center}
\centerline{\includegraphics[width=0.6\columnwidth]{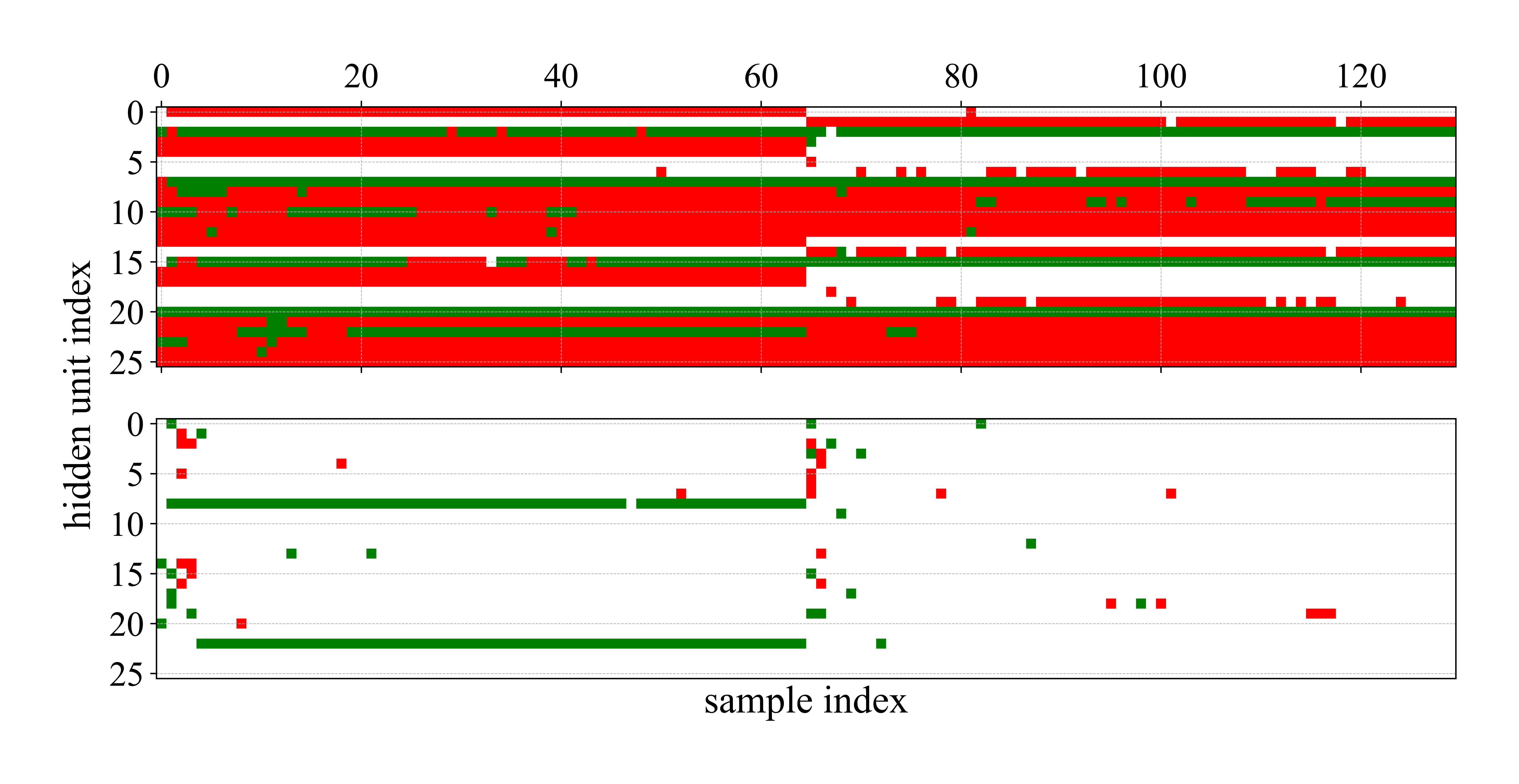}}
\vspace{-0.3cm}
\caption{Spatio-temporal spiking patterns for hidden circuits when presented with two MNIST-DVS samples, each of $65$ samples with green and red corresponding to the first and second spiking unit of a circuit, respectively. Top: Learning rule without regularization. Bottom: {\VOWEL} with sparsity level $r=0.3$ and $\alpha = 1$.}
\label{fig:reg}
\end{center}
\vspace{-0.3cm}
\end{figure}

{\bf Impact of variational regularization.} 
We now evaluate the impact of variational regularization on the performance of {\VOWEL}. To this end, we train a WTA-SNN with $H=16$ hidden circuits based on $3$ trials with different random seeds. In Fig.~\ref{fig:reg}, we observe the spiking behavior of hidden circuits in a WTA-SNN trained without (top) and with (bottom) regularization for two 65 time-steps long examples. While a WTA-SNN trained without regularization outputs dense spiking signals for the hidden circuits, the regularization term applied by {\VOWEL} clearly enforces temporal sparsity on the output of the hidden circuits. 

{\bf Comparison of the performance between {\VOWEL} and {\DECOLLE} on MNIST-DVS dataset.} 
The training dataset is composed of 900 examples per class, and the test dataset is composed of 100 samples per class. As is done for the DVSGesture dataset in the main text, we propose a comparison of the performance between {\VOWEL} and {\DECOLLE} in Table~\ref{tab:comparison-mnist}. Table~\ref{tab:comparison-mnist} is organized as Table~\ref{tab:comparison-gesture}. While the performance of {\DECOLLE} is quite sensitive to the number of hidden neurons, for this simpler task the degradation is not as dramatic as for DVSGesture. We attribute this result to the fact that MNIST-DVS images can be successfully classified even with rate-encoded models \cite{kaiser2018decolle, zhao2014feedforward, henderson2015spike}, hence not requiring the additional discriminatory power of WTA-SNNs.  

\begin{table}[ht]
\caption{Test accuracy of {\VOWEL} and {\DECOLLE} on MNIST-DVS}
\label{tab:comparison-mnist}
\vskip 0.1in
\begin{center}
\begin{small}
\begin{sc}
\begin{tabular}{lcccr}
\toprule
Model & Period & $H$ & Accuracy \\
\midrule
{\DECOLLE} \cite{kaiser2018decolle} & 1 ms & 512 &$93.18 \pm 0.52 \%$  \\
 & 10 ms & $512$ & $92.20 \pm 1.09\%$  \\
 & 25 ms & $512$ & $92.42 \pm 0.08\%$ \\
 & 25 ms & $256$ & $89.08 \pm 0.14\%$ \\
 & 25 ms & $128$ & $86.28 \pm 0.64\%$ \\
 & 25 ms & $64$ & $79.62 \pm 2.31\%$ \\

\midrule
& 25 ms & 256 & $84.47 \pm 0.42\%$ \\
{\VOWEL} & 25 ms & 128 & $82.06 \pm 0.66 \%$ \\
 & 25 ms & 64 & $80.17 \pm 0.45 \%$ \\ 
\bottomrule
\end{tabular}
\end{sc}
\end{small}
\end{center}
\vspace{-0.5cm}
\end{table}


\subsection{DVSGesture}
This dataset contains a set of 11 hand gestures performed by 29 subjects under 3 illumination conditions, which are captured in a similar fashion to the MNIST-DVS dataset using a DVS128 camera. 
Recordings are stored as vectors of one-hot encoded pixels, taking into account the polarity of events. Following the preprocessing steps taken in \cite{kaiser2018decolle}, we downsized the images to $32 \times 32$ pixels by summing the events in four neighboring pixels, and downsampled by binning in frames of variable length. During training, 500 ms long sequences are presented for each sample, while testing is performed on sequences of $1800$ ms.

\subsection{Hyperparameter selection}
Regularization in \eqref{eq:wta-ls} assumes an i.i.d. reference categorical distribution with a desired spiking rate $r$, i.e., $\log r(\bmh_{\leq T}) = \sum_{t=1}^T \sum_{i \in \set{H}} \sum_{c=1}^C {\bm 1}_{\{\bmh_{i,t} = \bfe_c\}} \cdot \log (r/C) + {\bm 1}_{\{\bmh_{i,t} = {\bm 0}\}} \cdot \log (1-r)$. During testing, the predicted class is selected as the index of the output circuit in the read-out layer with the largest overall number of output spikes across all units. Hyperparameters have been selected after a non-exhaustive manual search and are mostly shared among experiments. The learning rate is halved after each epoch. We note that a more extensive selection could lead to potential accuracy gains. Values of the hyperparameters are summarized in Table~\ref{tab:hyperparameters}.

\begin{table}[ht]
\caption{Hyperparameters used for {\VOWEL}}
\label{tab:hyperparameters}
\vskip 0.15in
\begin{center}
\begin{small}
\begin{sc}
\begin{tabular}{ccc}
\toprule
Parameter & Description & Value \\
\midrule
$K$ & Number of spatio-temporal filters & $8$ (MNIST-DVS) / $10$ (DVS Gesture) \\
$\tau$ & Filters duration & $10$ \\ 
$\eta$ & Learning rate & $0.05 / H$ \\
$\alpha$ & KL regularization factor & $1$ \\
$\gamma, \kappa$ & temporal averaging factor & $0.2$ \\
$\kappa_b$ & Baseline averaging factor & $0.05$ \\
$r$ & Desired spiking sparsity of hidden neurons & $0.3$ \\ 
\bottomrule
\end{tabular}
\end{sc}
\end{small}
\end{center}
\vspace{-0.3cm}
\end{table}


\end{document}